\pdfoutput=1

\documentclass[11pt]{article}

\usepackage{acl}

\usepackage{times}
\usepackage{latexsym}

\usepackage[T1]{fontenc}

\usepackage[utf8]{inputenc}

\usepackage{microtype}

\usepackage{inconsolata}
\usepackage{url}
\usepackage{graphicx}%
\usepackage{subfigure}
\usepackage{booktabs} 
\usepackage{siunitx}

\usepackage{booktabs}
\usepackage{caption}
\usepackage{colortbl}
\usepackage{multirow}

\usepackage{rotating}
\usepackage{wrapfig}

\usepackage{amsmath,amsfonts,bm}
\definecolor{darkblue}{RGB}{25, 50, 112}
\definecolor{ROW_COLOR}{HTML}{C9F7F4}
\definecolor{COLOR_MEAN}{HTML}{f0f0f0}
\usepackage{url}

\title{
Unlocking Emergent Modularity in Large Language Models 
}

\author{$^{1}$$^{2}$Zihan Qiu\thanks{Equal contribution}\thanks{Work done while interning at HKUST.} $^{ }$,  $^{3}$Zeyu Huang$^*$,  $^{1}$Jie Fu\thanks{Corresponding author}  \\
$^{1}$CSE, HKUST $\,\,\,$ $^{2}$IIIS, Tsinghua University, \\
$^{3}$ILCC, University of Edinburgh  \\
\texttt{qzh11628@gmail.com, zeyu.huang@ed.ac.uk, jiefu@ust.hk}}

\begin{document}
\maketitle
\begin{abstract}
Modular Neural Networks (MNNs) demonstrate various advantages over monolithic models.
Existing MNNs are generally \textit{explicit}: 
their modular architectures are pre-defined, with individual modules expected to implement distinct functions.
Recent works reveal that there exists \textit{implicit} modularity in standard pre-trained transformers, namely \textit{Emergent Modularity}.
They indicate that such modular structures spontaneously exhibit during the early pre-training phase.
Despite the benefits of modularity, most Language Models (LMs) are still treated as monolithic models in the pre-train and fine-tune paradigm, with their emergent modularity locked and underutilized.
In this work, focusing on unlocking the emergent modularity in LMs, we showcase that standard LMs could be fine-tuned as their Mixture-of-Expert (MoEs) counterparts without introducing any extra parameters. 
Such MoEs are derived from emergent modularity and are referred to as Emergent MoEs (EMoE).
Our experiments demonstrate that fine-tuning EMoE effectively improves downstream in-domain and out-of-domain generalization compared with vanilla fine-tuning.
Our analysis and ablation studies further illustrate that it is robust to various configurations and can scale up to Large Language Models (i.e., Llama2-7B and Llama-30B). Code is available at \href{https://github.com/qiuzh20/EMoE}{this repo}.
\end{abstract}

\section{Introduction}
Modularity attracts considerable attention from the artificial intelligence community~\citep{DBLP:journals/ijns/AudaK99}.
Neural networks with modular designs, termed Modular Neural Networks (MNNs), have exhibited a wide range of advantages, including adaptation~\citep{DBLP:journals/corr/abs-2306-04640}, data efficiency~\citep{DBLP:conf/iclr/BengioDRKLBGP20}, and better generalization abilities~\citep{DBLP:journals/corr/abs-2011-15091, DBLP:conf/nips/WeissRLPBSLB22}. 
Typical MNNs are usually \textit{explicitly} modular.
They have a pre-defined modular structure and are expected to achieve a divide-and-conquer solution for the given task.
Among various MNNs, Mixture-of-Experts (MoEs) employ a conditional computation strategy where different submodules - so-called experts - are expected to be activated by different types of inputs. 
MoEs see substantial success in various domains~\citep{
DBLP:journals/jmlr/FedusZS22, 
DBLP:journals/corr/abs-2303-07226, DBLP:conf/cvpr/ChenSDCZLG23,DBLP:conf/nips/MustafaRPJH22, DBLP:conf/nips/BaoW0LMASPW22} in the era of large-scale transformers,
and they are therefore a widespread modular neural architecture.

Apart from \textit{explicit} MNNs, some research finds that modular structures spontaneously emerge during training, not only in small-scale CNNs or LSTMs~\citep{DBLP:conf/iclr/CsordasSS21, DBLP:conf/iclr/AgarwalaDJPSWZ21}, but also large-scale pre-trained transformer models.
\citet{DBLP:conf/acl/ZhangL00S022, DBLP:journals/corr/abs-2210-06313} reveal notable sparse activation patterns within the Feed-Forward Networks (FFNs) in pre-trained transformer models. They find that in T5-Base~\citep{DBLP:journals/jmlr/RaffelSRLNMZLL20} and ViT-B16, only 3.0\% and 6.3\% neurons are activated during one forward process, respectively.
Furthermore, ~\citet{DBLP:conf/acl/ZhangZLXW00XS023} utilize handpicked semantic and knowledge-intensive tasks to probe the nature of neurons in FFNs.
They observe a strong correlation between neuron activation and specific tasks, further discovering clear function-based neuron grouping of the pre-trained T5 model (neurons with similar functions are usually co-activated). They summarize such phenomenon as \textit{Emergent Modularity} (EM).

Though modularity emerges, pre-trained language models are generally treated as monolithic models in the standard pre-train and fine-tune paradigm. 
It is natural to ask \textit{whether their EM and the potential improvements brought by EM are locked in this process.}

In this paper, we advocate unlocking the EM in the pre-trained language models could bring generalization improvements for downstream tasks.
Specifically, we split certain FFNs layers of the original model into MoEs layers. 
The MoEs is derived according to the EM in that layer and can be regarded as the externalization of the EM. Hence, the obtained MoEs model is called Emergent MoEs (EMoE).
We then fine-tune the obtained EMoE model to investigate whether unlocking EM encourages downstream task performance.

We validate our empirical findings with various models, evaluation benchmarks, fine-tuning methods (parameter-efficient tuning and full fine-tuning). 
We find that fine-tuning EMoE achieves stronger generalization performance than vanilla fine-tuning across various experimental settings, 
demonstrating that unlocking the EM of LMs boosts the models' downstream generalization abilities.
We provide a comprehensive analysis for EMoE: 1) We first validate that EMoE indeed unlocks EM in pre-trained language models by showcasing its task-specific expert choice.
2) We then reveal that EMoE ameliorates the parameter updating during fine-tuning and can even be abandoned afterward.
We want to highlight that this property further improves the practicality of EMoE as the model architecture does not change before and after fine-tuning. Meanwhile, our ablation studies show the EMoE's robustness to various hyper-parameter configurations. 
3) We also conclude that EMoE could mask neurons with negative transfer effects.
We hope our research discoveries can bring novel insights and serve as an example attempt towards further unlocking the EM of LLMs.

\section{Methodology}
\subsection{Preliminaries}
\textbf{Transformer FFNs are Key-Value Memories.} The FFNs layer in the transformer block typically includes weights $\mathbf{K} \in \mathbb{R}^{h\times d}$, 
$\mathbf{V} \in \mathbb{R}^{d\times h}$, 
where $h$ is embedding size and $d$ is the dimension of the hidden layer (usually $d=4h$), and a non-linear activation function $\sigma(\cdot)$. 
For an input $\mathbf{x}\in\mathbb{R}^{h}$, the output $\mathbf{y}\in\mathbb{R}^{h}$ can be calculated as Eq.~\ref{eqn:FFN_0}:
\begin{equation}
\label{eqn:FFN_0}
\mathbf{y}=\operatorname{FFN}(\mathbf{x};\mathbf{K},\mathbf{V})=\sigma(\mathbf{x}\cdot \mathbf{K})\cdot \mathbf{V}.
\end{equation}
More precisely, for each column $\mathbf{K}_{:, i}$ and row $\mathbf{V}_{i, :}$, Eq.\ref{eqn:FFN_0} can be rewritten as:
\begin{equation}
\label{eqn:FFN_1}
\mathbf{y}=\sigma(\mathbf{x}\cdot \mathbf{K})\cdot \mathbf{V}=\sum_{i=1}^{h}\sigma(\mathbf{x}\cdot\mathbf{K}_{:, i})\cdot \mathbf{V}_{i, :}
\end{equation}

Following \citet{DBLP:conf/emnlp/GevaSBL21, DBLP:conf/emnlp/GevaCWG22,DBLP:conf/iclr/HuangSZZR023}, we regard columns in $\mathbf{K}$ as key vectors and rows in $\mathbf{V}$ as value vectors, the output of an FFNs network can be viewed as a weighted sum of value vectors based on the activation scores $\sigma(\mathbf{x}\cdot \mathbf{K})$.
For the rest of this paper, we refer to one key-value memory pair using \textit{neuron} and the co-activation of neurons using \textit{modularity}.

\textbf{Mixture-of-Experts} 
In transformers, MoEs is often implemented by replacing the original FFNs with a group of parallel FFNs and introducing a gating module. 
Supposing there are $N$ experts: $\{
\operatorname{FFN}^n(\cdot;\mathbf{K}^n,\mathbf{V}^n)|\,n\in[1, N]\}$, 
the gating module $g(\cdot\,;\mathbf{G},k)$, 
defined with its parameters $\mathbf{G}$ and an integer $k$, is to map input $\mathbf{x}$ to a score distribution of experts $g(\mathbf{x};\,\mathbf{G},k) \in \mathcal{R}^N$. 
Typically, $g$ is implemented with a simple linear layer followed by a $\operatorname{softmax}$ function and a $\operatorname{Top-k}$ function.
Given $\mathbf{x} \in\mathbb{R}^{h}$, 
the output $\mathbf{y}\in \mathbb{R}^{h}$ of can be summarized as the weighted sum of the output from all experts: 
\begin{equation}
\label{eqn:MoE}
\mathbf{y}=\sum_{n\in N}g_n(\mathbf{x};\mathbf{G},k)\operatorname{FFN}^n(\mathbf{x};\mathbf{K}^n,\mathbf{V}^n)
\end{equation}
When $k$ for $\operatorname{Top-K}$ is smaller than $N$, only a subgroup of experts is involved in the computation, termed sparse MoEs.

\subsection{Emergent Mixture-of-Expert}
\label{sec:decompose_method}
\begin{figure*}[t!]
    \vskip  -0.3in
  \begin{center}\centerline{\includegraphics[width=\textwidth]{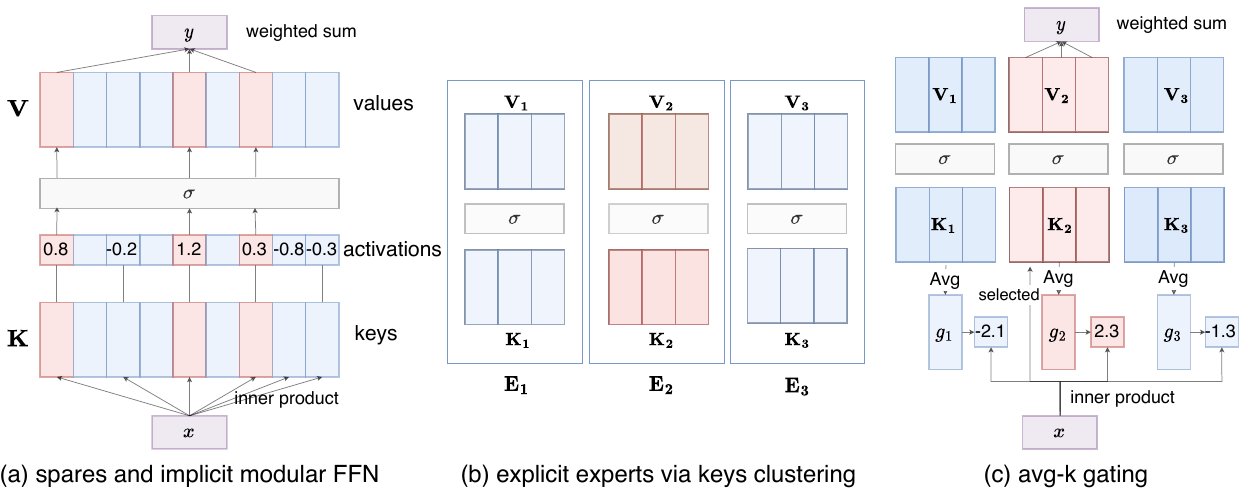}}
  \caption{
  (a) Existing literatures~\citep{DBLP:conf/emnlp/GevaSBL21, DBLP:conf/emnlp/GevaCWG22} suggest that the FFNs in transformers can be viewed as key-value memories. They regarded the input as a query, the first layer as keys, and the second as values. 
  Given an input, keys are sparsely activated (marked in \textcolor{red}{red}). Most of the values don't impact the output. 
 (2) The FFNs block can be partitioned into experts by clustering keys.
 (3) Afterward, experts' key averages are used as the gating weights. The inner product between $\textbf{x}$ and gating weights are used to select experts.
  }
  \label{fig:Emoe}
  \end{center}
  \vskip  -0.3in
\end{figure*}

According to Eq.~\ref{eqn:FFN_1} and Eq.~\ref{eqn:MoE}, we find that FFNs internally resemble MoEs if we consider keys as the gating module and values as the group of experts, which inspire us to transform existing FFNs into sparse MoEs to unlock its modular potentials.
Since our research goals mainly focus on EM, a preferred approach to externalizing EM into sparse MoEs should not introduce additional parameters, training, and data, which may result in impractical or undesired biases. 
Therefore, after splitting original \textit{neurons} into different groups to construct different experts, each group's average of key vectors is calculated to form the gating module. Details are described and illustrated in Fig.~\ref{fig:Emoe}.

\textbf{Clustering-based Experts Construction.}
We aim to ensure that \textit{neurons} that tend to be co-activated are divided into the same group.
Since \textit{neurons} with similar key vectors tend to be co-activated according to Eq.~\ref{eqn:FFN_1}, we split them into separate experts by clustering their key vectors. 
Specifically, given a FFNs layer $\operatorname{FFN}(\cdot;\mathbf{K},\mathbf{V})$, we perform constrained clustering~\citep{DBLP:conf/sspr/MalinenF14} (more details are in Appendix~\ref{appendix:constrained})  to partition all key vectors $\mathbf{K}$ into $N$ experts on average. 
Denoting the indices of keys in the $i$-th group as $E_i \subset [d]$, for $\forall j \in E_i$, we extract \textit{neuron} $(\mathbf{K}_{:,j}, \mathbf{V}_{j,:})$ to form the $i$-th expert $\operatorname{FFN}(\cdot;\mathbf{K}^i,\mathbf{V}^i)$ as depicted in Fig.~\ref{fig:Emoe}(b).
After that, the computation of each expert follows Eq.~\ref{eqn:FFN_0}.

\textbf{Avg-k Gating.} 
As discussed, we do not want to introduce extra trainable parameters when externalizing EM.  
Therefore, we construct the gating module by averaging each expert's keys, which should route the input $\mathbf{x}$ to the experts who tend to have larger activation scores and thus contribute more to the model's output.
The gating function is usually implemented by a single layer $\mathbf{G}\in\mathcal{R}^{h\times N}$, in the $\operatorname{avg-k}$ gating's case, the weights in $i$-th column $\mathbf{G}_{:,i}$ can be calculated as follows:
\begin{equation}
\label{eqn:expert_embeddings}
\mathbf{G}_{:,i}=\operatorname{Avg}(\mathbf{K}^i,\text{dim=0}).
\end{equation}
And then the gating score for $i$-th expert is:
\begin{equation}
\label{eqn:avg_k_gate}
g_i(\mathbf{x};\mathbf{G},k)=
\begin{cases}
&1~~\text{ if } i \in \operatorname{Top-K}(\mathbf{x}\cdot \mathbf{G};k) \\ 
&0~~\text{ else}
\end{cases}
\end{equation}
where $\operatorname{Top-K}(\cdot;k)$ returns indices of $k$ largest element of the given input along.
As the gating score is the average of activation scores (before $\sigma(\cdot)$) of neurons in that expert:
\begin{equation}
\label{eqn:avg_k_gate_and_expert}
\mathbf{x} \cdot \mathbf{G}_{:,i} = 
\mathbf{x} \cdot\operatorname{Avg}(\mathbf{K}^i) = \frac{N}{d} \sum_{j\in E_i}\mathbf{x} \mathbf{K}^i_{:,j}  = \frac{N}{d} \sum_j a_j.
\end{equation}
a larger value of gating score $g_i$ implies more activated keys within the corresponding expert. Consequently, the expert could potentially contribute more to the output $\mathbf{y}$ for input $\mathbf{x}$. 
During downstream tuning, gating weights are tied with the FFNs parameters using Eq.~\ref{eqn:expert_embeddings}.

\section{Experiments}
\label{sec:main_results}
\textbf{Configurations}: 
We first evaluate EMoE using BERT and GPT2 series models.
We employ GLUE~\citep{DBLP:conf/iclr/WangSMHLB19} and
GLUE-X~\citep{DBLP:conf/acl/YangZ0LWLW0023} for benchmarking in-domain (ID) and out-of-domain (OOD) performance of the fine-tuned model, respectively.
We mainly present the experimental results when employing LoRA~\citep{DBLP:conf/iclr/HuSWALWWC22} to fine-tune the pre-trained language models for two reasons: 
(1) with the increasing scale of pre-trained models, parameter-efficient tuning~\citep{DBLP:conf/icml/HoulsbyGJMLGAG19} can scale up to very large language models and thus becomes more practical.
(2) standard LoRA weights are added in each self-attention block, and the parameters in FFNs will not be updated, allowing us to investigate whether leveraging EM, even without fine-tuning the parameters of EMoE, can yield improvements.
We present EMoE full fine-tuning results in the Appendix~\ref{appendix:full-ft}.
Besides, we scale up EMoE on Llama2-7B and Llama-30B~\citep{DBLP:journals/corr/abs-2302-13971} for further validation. 
We instruct-tuning models on Alpaca and testing them on the MMLU benchmark~\citep{DBLP:conf/iclr/HendrycksBBZMSS21}.
For more details about datasets, evaluation metrics and computation cost, please refer to the Appendix~\ref{appendix:datasets_eval} and~\ref{appendix:computation_and_memory}.

\begin{table*}[tbp!]
\vskip  -0.3in
\small
    \setlength{\tabcolsep}{4pt}
    \renewcommand{\arraystretch}{1.0}
    \centering
    \caption{
    ID and OOD performance of EMoE and baseline models.
    All the reported results are obtained from 3 independent experiments. 
    OOD Metrics (averaged over 14 OOD tasks, lower is better) provide additional information for OOD generalization. The best result is highlighted in \textbf{bold}.}
    \vskip -0.07in
    \label{tab:results_lora_ID_OOD}
    \resizebox{0.9\textwidth}{!}{
    \begin{tabular}{l|ccccccccc|c}
    \toprule
    \rowcolor{COLOR_MEAN}
    \textbf{Algorithm} & \textbf{MRPC} & \textbf{CoLA} & \textbf{RTE} & \textbf{STSB} & \textbf{SST2} & \textbf{QNLI} & \textbf{QQP} & \textbf{MNLI} & \textbf{ID-Avg}$\uparrow$ & \textbf{$\scalebox{0.8}{OOD$\downarrow$}$} \\
    \midrule
    \multicolumn{11}{c}{BERT-Large (340M Parameters)} \\
    \midrule
    LoRA & 89.97 & 63.40 & 72.92 & \textbf{90.51} & 93.16 & 92.20  & 87.21 & 85.40 & 84.35 & 4.86 \\
    GMoE & 89.45 & 63.80 & 72.56 & 90.29 & \textbf{93.85} & 92.32 & \textbf{87.99} & \textbf{85.92} & 84.52(+0.18)& \textbf{4.04} \\
    $\scalebox{0.8}{EMoE-learn}$ & 89.87 & 64.00 & 71.36 & 90.48 & 93.65 & \textbf{92.40} & 87.55 & 85.62 & 84.37(+0.02)& 4.66 \\
     \rowcolor{ROW_COLOR}
    EMoE & \textbf{90.85} & \textbf{65.33} & \textbf{75.21} & 90.43 & 93.50 & 92.23 & 87.74 & 85.43 & \textbf{85.09(+0.74)}& 4.37 \\
    \midrule
    \multicolumn{11}{c}{GPT2-XL (1.5B Parameters)} \\
    \midrule
    LoRA & 86.83 & 60.88 & 78.70 & 89.07 & 95.18 & 91.84 & 87.41 & 86.93 & 84.61 & 5.61 \\
    GMoE & 87.02 & \textbf{62.81} & 79.78 & 89.21 & 95.41 & 92.18 & 89.10 & \textbf{87.17} & 85.34(+0.73)& 4.33 \\
    $\scalebox{0.8}{EMoE-learn}$ & \textbf{87.93} & 61.50 & 79.90 & \textbf{89.48} & 95.18 & \textbf{92.33} & \textbf{89.71} & 87.00 & 85.38(+0.77)& 4.40 \\
     \rowcolor{ROW_COLOR}
    EMoE & 87.75 & 62.27 & 80.02 & 89.37 & \textbf{95.41} & 92.10 & 89.58 & 87.06 & \textbf{85.45(+0.84)}& \textbf{3.88} \\
    \bottomrule
    \end{tabular}
    }
    \vskip -0.1in
\end{table*}

\textbf{Baselines} 
Our baselines include 
(1) \textit{vanilla LoRA-tuning}: add LoRA weights to the q and v projections in attention layers; 
(2) \textit{GMoE}~\citep{DBLP:conf/iclr/LiSYWRCZ023}: Instead of splitting, GMoE replicates FFNs layer and train new gating layers to introduce MoE structure from the two-to-last and fourth-to-last transformer blocks in the original pre-trained models. 
Since GMoE copies the FFNs of the pre-trained model to form the MoEs, it is ineffective if introduced MoEs aren't tuned.
Thus, we conduct experiments with LoRA tuning for GMoE and tune the transformer block where the original FFNs are replaced.
(3) \textit{EMoE-learn}: an ablation method, where the gating function is learned (same as GMoE) during fine-tuning. 
This helps us better understand the effect of $\operatorname{avg-k}$ gating.

\textbf{Hyper-parameters}: those unrelated to MoEs (e.g., learning rate, batch size) remain consistent with the baselines. Following~\citet{DBLP:conf/iclr/LiSYWRCZ023}, we replace original FFNs with EMoE layer in \{last two even layers, last one even layer\}.
Comparable hyper-parameter searches are conducted for both GMoE and EMoE for the number of experts $N$ and top-k: GMoE explores $N$ within \{4, 8\} and top-k within \{1, 2\}; for EMoE, $N$ is fixed at 64, with top-k explored within \{16, 32\}.
The underlying reason for different $N$ and top-k in GMoE and EMoE is that EMoE decomposes the original FFNs into experts, and the total parameters of all experts remain the same as the original FFNs. 
On the contrary, GMoE replicates the original FFN from a MoEs architecture; the larger $N$ and top-k they choose, the more parameters and computational overhead they introduce.
Our ablation study indicates that while more careful hyper-parameter searches may yield superior performance, adhering to a $\text{top-k}/N=0.25 \text{ or } 0.5$ for EMoE consistently brings improvement over vanilla fine-tuning.

\textbf{Evaluation Metrics}
All experiments except the scale-up ones are repeated three times, and the average is presented in the main section. Full results are in Appendix~\ref{appdendix:full_tables}. 
For the \textit{OOD metric}, we follow GLUE-X~\citep{DBLP:conf/acl/YangZ0LWLW0023} and employ the Friedman rank~\citep{friedman1940comparison}\footnote{Our evaluation includes nine methods detailed, demonstrating the relationships among these methods. Therefore, these values can’t provide a direct comparison with the results in the GLUE-X paper} $\text{rank}_f=\frac{1}{n}\sum_{i=1}^n \text{rank}_i$. 
For each method under the same backbone, $\text{rank}_i$ is produced based on each dataset's best and average results. 
For the 13 OOD tasks used in GLUE-X, each method generates 26 $\text{rank}_i$ values. 
The OOD results presented in Table~\ref{tab:results_lora_ID_OOD} represent the mean of all these $\text{rank}_i$ values.
The original results of each task can be found in the Appendix~\ref {appdendix:full_tables}.

\textbf{Results with BERT and GPT2}
According to Tab.~\ref{tab:results_lora_ID_OOD}: 
(1) EMoE demonstrates enhancements compared to vanilla LoRA-tuning.
Notably, EMoE also achieves results comparable to GMoE on BERT-large and outperforms GPT2-XL with much fewer learnable parameters. 
(2) While EMoE-learn's results are better than EMoE in several tasks (STSB, QNLI), EMoE exhibits higher stability than EMoE-learn and delivers superior overall results.
(3) MoEs structures improve OOD performance (GMoE, EMoE, EMoE-learn vs. LoRA). 

\begin{table}[h]
\centering
\caption{Instruction LoRA-tuned Llama models evaluated (in accuracy) on the MMLU benchmark. Times are wall-clock computation times for training 3 epochs on Alpaca using the same GPU devices.
When $N=K=1$, it represents the vanilla LoRA tuning.
}
\vskip  -0.1in
\label{tab:llama-lora}
\resizebox{0.5\textwidth}{!}{
\begin{tabular}{c|c|c|c|c}
\toprule
 Experts (N) & topk (k) & MMLU score & times (s) & FLOPS ($10^{17}$) \\
\midrule
\multicolumn{5}{c}{Llama2-7B} \\
\midrule
1 & 1 & 46.96 & 1396 & 6.92 \\
64 & 16 & \textbf{47.58} & 1545 & 7.03 \\
64 & 32 & 47.37 & 1521 & 7.13 \\
\midrule
\multicolumn{5}{c}{Llama-30B} \\
\midrule
1 & 1 & 56.18 & 6943 & 22.5 \\
256 & 64 & \textbf{57.11} & 6955 & 22.5 \\
256 & 128 & 56.64 & 6974 & 22.4 \\
\bottomrule
\end{tabular}
}
\vskip  -0.1in
\end{table}

\begin{figure*}[!t]
\vskip  -0.3in
\begin{center}
\includegraphics[width=0.62\columnwidth]{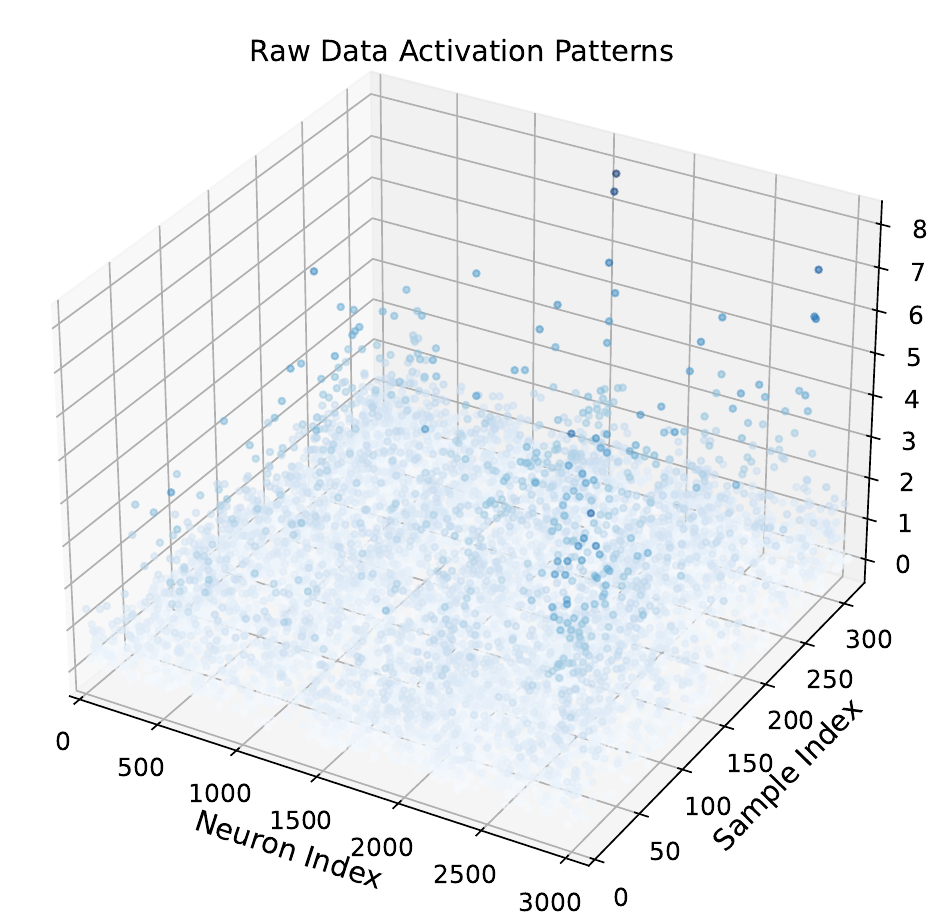}
\includegraphics[width=0.62\columnwidth]{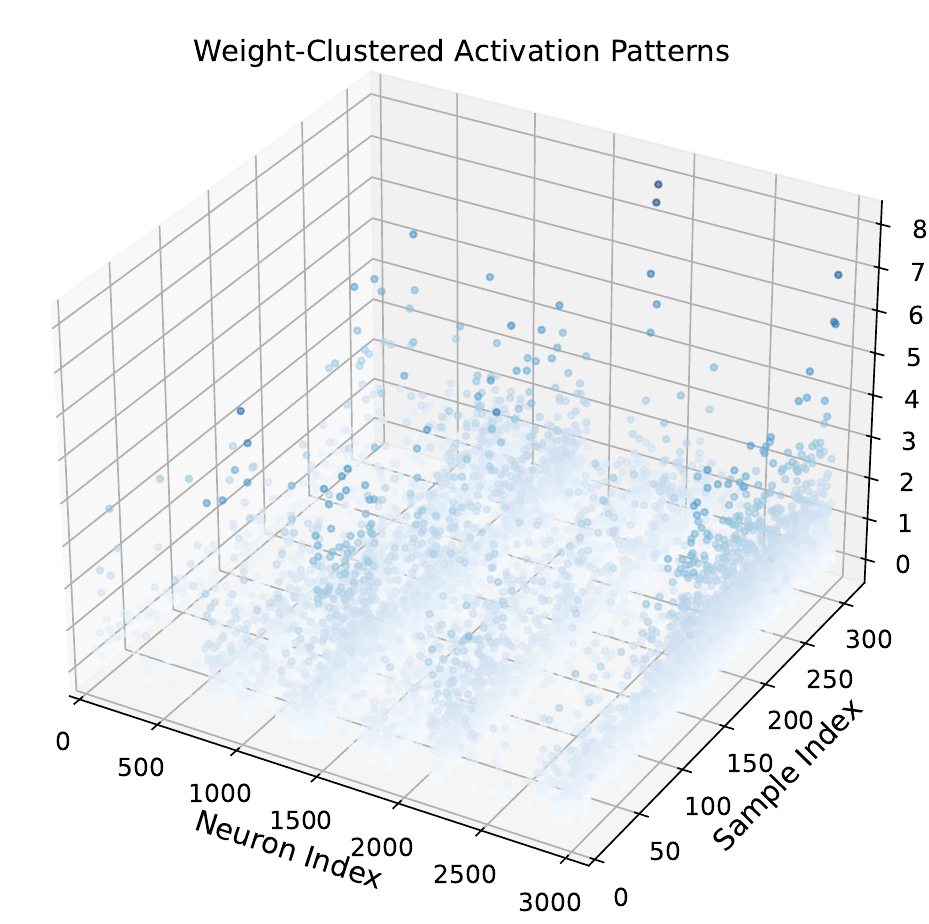}
\includegraphics[width=0.72\columnwidth]{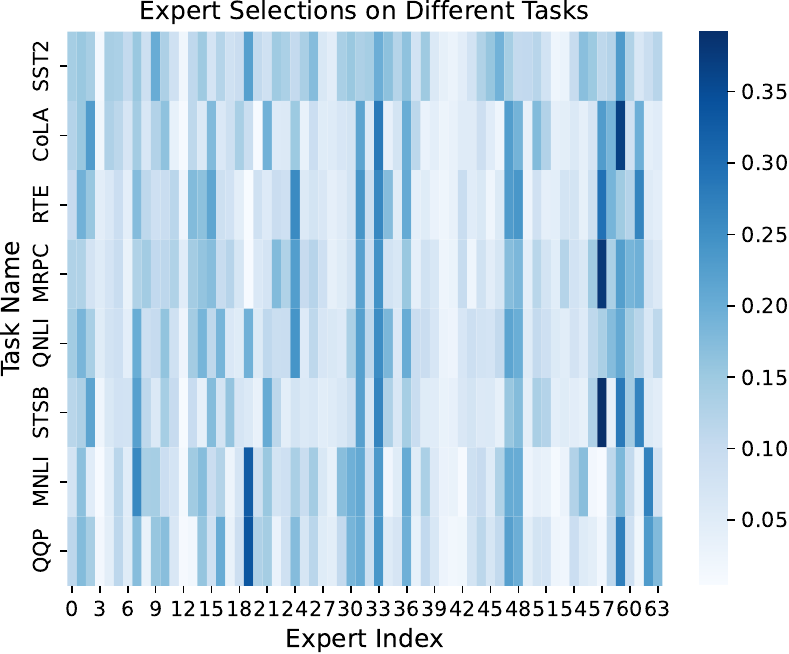}
\caption{
\textbf{Left}: Activations of neurons (z-axis denotes activation value) in FFNs of a pre-trained transformer models.
\textbf{Middle}: By clustering the keys in the FFNs layer and rearranging the activation scores accordingly, modular patterns of neuron activation emerge.
\textbf{Right}: The heat map between experts and tasks. It is observed that the activation of experts is task-dependent, while some experts are generally shared across different tasks.
}
\label{fig:clustering}
\end{center}
\vskip  -0.2in
\end{figure*}

\textbf{Results with Llama}
Unlike GMoE, EMoE does not introduce any additional trainable parameters or procedures so that we can scale EMoE up to models of 7B and 30B sizes to validate its effectiveness further.
From Tab.~\ref{tab:llama-lora}:
1. EMoE consistently yields improvements compared with LoRA tuning with negligible additional computation.
2. Because EMoE does not introduce any additional trainable parameters, the choice of K and N proportions remains large (can be even larger) and is applicable in larger-scale models. While the performance varies when employing different $N$ and $K$, EMoE consistently outperforms the vanilla LoRA tuning.
This suggests that although additional hyperparameters $N$ and $K$ are introduced, they do not lead to usability challenges. Please refer to Sec.~\ref{sec:Analysis_Ablation} for more ablation results on hyperparameters.

\textbf{Additional Results}
Moreover, we conduct experiments with full fine-tuning under comprehensive evaluation settings:
1. Vision OOD benchmark Domainbed~\citep{DBLP:conf/iclr/GulrajaniL21}, full fine-tuning ViT-Small (22M) and ViT-Base (86M).
2. GLUE benchmark, fully fine-tuning BERT-Base, BERT-Large, and GPT2-Small.
3. Full fine-tuning Llama2-7B.
We present detailed results and analyses in Appendix~\ref{appendix:full-ft}. 
According to the results, EMoE brings consistent improvements over standard full finetuning across various tasks and model scales. 
On the vision OOD benchmark Domainbed, which strictly controls evaluation metrics, EMoE achieved results comparable to the state-of-the-art GMoE. Furthermore, when full-finetuned with Alpaca, EMoE exhibited a notable improvement of \textbf{1.58} on MMLU to the standard tuning baseline.
These findings underscore the effectiveness of EMoE in enhancing model performance across various architectures and tasks.

\section{Analysis}
\label{sec:analysis}

\subsection{Does EMoE Unlock Emergent Modularity?}

We first investigate whether simple key-vector-based clustering partitioning could capture the modular pattern of the neuron activation.
The activation scores of different neurons on different inputs are visually shown in the Fig.~\ref{fig:clustering} left and middle. 
Before clustering (Fig.~\ref{fig:clustering} left), the activation of neurons seems random. After rearranging those activation scores according to the EMoE partition (Fig.~\ref{fig:clustering} middle), we observed that only a part of the neurons are frequently employed in this task, and clear activation clusters emerge. 
This demonstrates that \textbf{key-vector-based clustering can decompose modular components within the standard model}.
We then delve into expert usages across different tasks in the EMoE.
The heatmap between tasks and experts (Fig.~\ref{fig:clustering} right) showcases that the expert utilization differs between tasks, while some crucial experts are nearly used by all 8 tasks.
For example, expert 19, that heavily used in QQP and MNLI, is merely activated in MRPC and RTE. On the contrary, expert 33 is frequently activated by all 8 tasks.

\subsection{How does EMoE improve fine-tuning performance?}

Though EMoE achieves notable improvements in both ID and OOD scenarios, it is non-trivial that simply transforming the pre-trained FFNs into MoEs before fine-tuning can yield such benefits, especially when using LoRA tuning, where the gates and experts part are not updated. We investigate the mechanism behind the enhanced performance.
We use BERT-Large as the backbone model.

\textbf{EMoE benefits LoRA weight learning instead of influencing inference.} 
\label{para:Analysis_when}
EMoE and the standard model only differ in FFNs activations. Such differences might (1) directly impact outputs during inference and (2) influence the parameter updating during training.
In light of this, we propose two variants and compare them with vanilla LoRA tuning and EMoE: 
(a) LoRA2EMoE: Using LoRA to fine-tune the original model and split it into EMoEs during inference. 
If it surpasses the vanilla LoRA tuning, we can infer that EMoE mainly impacts model inference. 
(b) EMoE2LoRA: Using LoRA to fine-tune an EMoE model and merge experts into original FFNs during inference. 
If no changes occur, it implies that EMoE mainly ameliorates the parameter updating of the fine-tuning stage.

\begin{figure}[h!]
\begin{center}
\includegraphics[width=0.7\columnwidth]{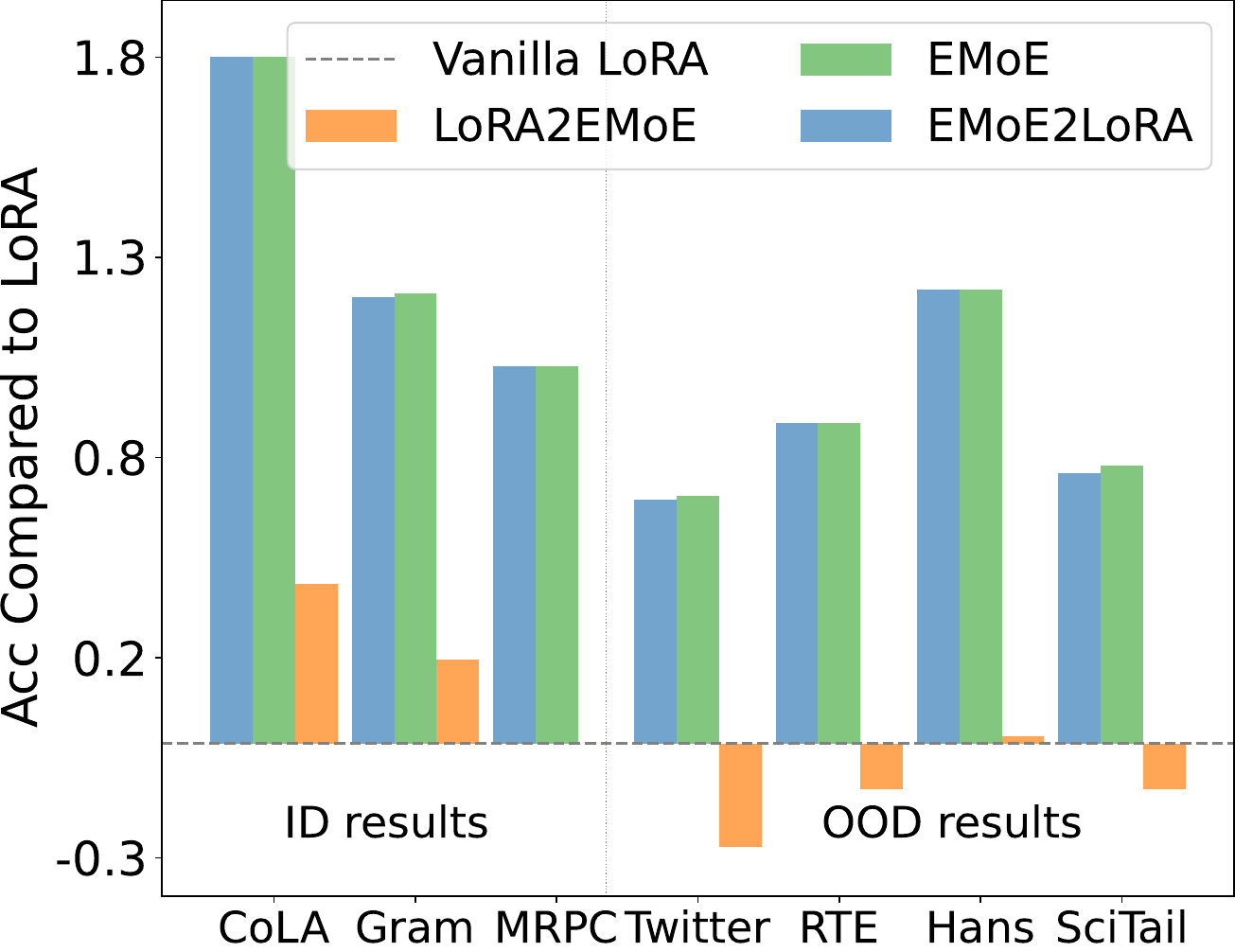}
\vskip -0.1in
\caption{ID and OOD accuracies compared with LoRA for validating EMoE's training \& inference effects.}
\label{fig:results-analysis}
\end{center}
\vskip -0.2in
\end{figure}

According to Fig.~\ref{fig:results-analysis}, doing sparse activation during testing does not contribute to better generalization on average (LoRA2EMoE). 
However, when merging EMoE back into the original FFNs after fine-tuning, the performance remains significantly better than the vanilla LoRA tuning and is almost identical to EMoE (EMoE2LoRA and EMoE).
Please refer to Appendix~\ref{appdendix:full_tables} Tab.~\ref{tab:results_compare_train_test_std} for full results.
Thanks to this property, we can use EMoE during fine-tuning and then convert the models to standard ones.
\textit{This allows the model to enjoy the benefits of EM without any alterations to its deployment, which improves EMoE's practicability in the LLMs era.}
We further validate this on the Llama series models and the findings are consistent. Thus, in Tab.~\ref{tab:llama-lora}, we report the results using EMoE to fine-tune and standard Llama architecture for benchmarking. And we highlight that this property can also be applied in the full fine-tuning setting, as illustrated in Appendix~\ref{appendix:full-ft} Tab.~\ref{tab:llama2-7B-ft}.

\begin{figure}[h!]
\begin{center}
\includegraphics[width=0.7\columnwidth]{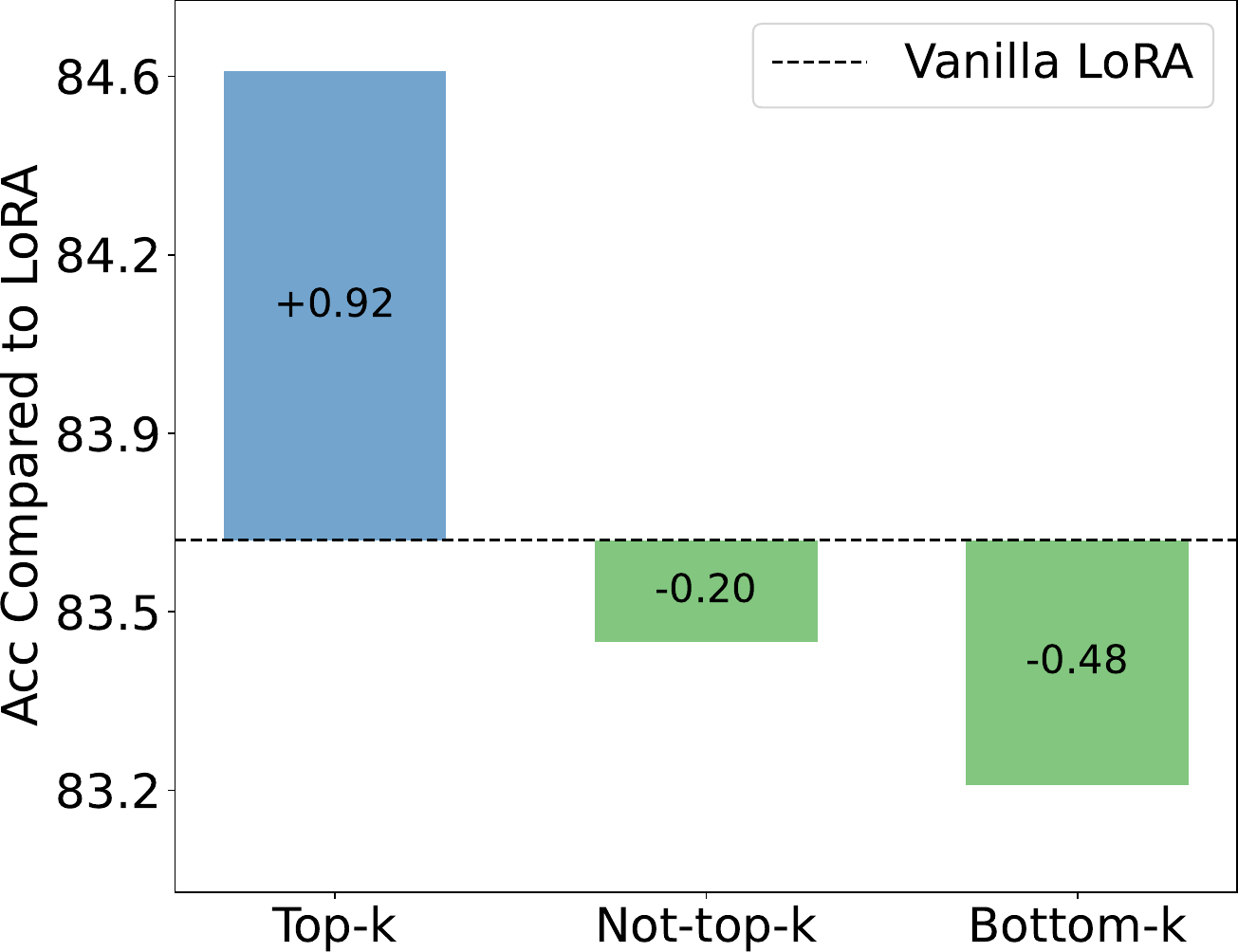}
\vskip -0.1in
\caption{Sparse activated training accuracies with different expert selections.}
\label{fig:results-neg}
\end{center}
\vskip -0.2in
\end{figure}

\textbf{EMoE masks neurons with negative transfer impacts.}
\label{para:Analysis_How_train}
The only difference between EMoE and vanilla LoRA tuning is EMoE blocks some activated neurons during training by $\operatorname{Top-k}$ expert selection.
Based on these, we hypothesize that the effects of EMoE stem from preventing negative knowledge transfer from blocked neurons.
Therefore, we investigate whether there are such negative transfers.
Specifically, we study the following expert selection variants: (1) Bottom-k: select $k$ experts with the \textit{lowest} scores; (2) Not-top-k: select $N-k$ experts who are not among the top-k experts.
These variants are evaluated across six tasks from GLUE. 
The averaged outcomes are in Fig.~\ref{fig:results-neg}. Full results can be found in Appendix~\ref{appdendix:full_tables} Tab.~\ref{tab:full_results_selections}.
LoRA tuning results with Bottom-k and Not-top-k expert selections are worse than vanilla LoRA tuning, while Top-K outperforms it.
We also examine the neuron activation ratio (the number of activated neurons in selected experts versus the number in the FFNs) across these variants. Full results can be found in Appendix~\ref{appdendix:full_analysis_results} Tab.~\ref{tab:activation_ratio}.
The activation ratios for these three variants are approximately 0.43 for Top-k, 0.57 for Not-top-k, and 0.12 for Bottom-k, respectively.
Notably, Not-top-k significantly lags behind Top-k, even though it involves and activates more neurons, suggesting the performance drop is more related to the neurons' property.
This further corroborates that masked neurons have negative transfer effects.

\begin{table}[ht]
\centering
\caption{T5-base Multi-tasks ID and OOD accuracies across different EMoE settings.}
\vskip -0.1in
\resizebox{0.4\textwidth}{!}{\begin{tabular}{cc|c|c}
\toprule
Experts ($N$) & topk ($k$) &  ID avg & OOD avg \\
\midrule
Baseline & - &  65.73 &  51.65\\
8 & 2 &  67.98 & 52.19 \\
16 & 4 & 68.14 & \textbf{53.23}\\
32 & 8 & \textbf{73.29} & 53.20\\
\bottomrule
\end{tabular}
}
\vskip -0.1in
\label{tab:T5-results}
\end{table}

\begin{figure*}[t!]
\vskip -0.5in
\begin{center}
\centerline{
\includegraphics[width=1.7\columnwidth]{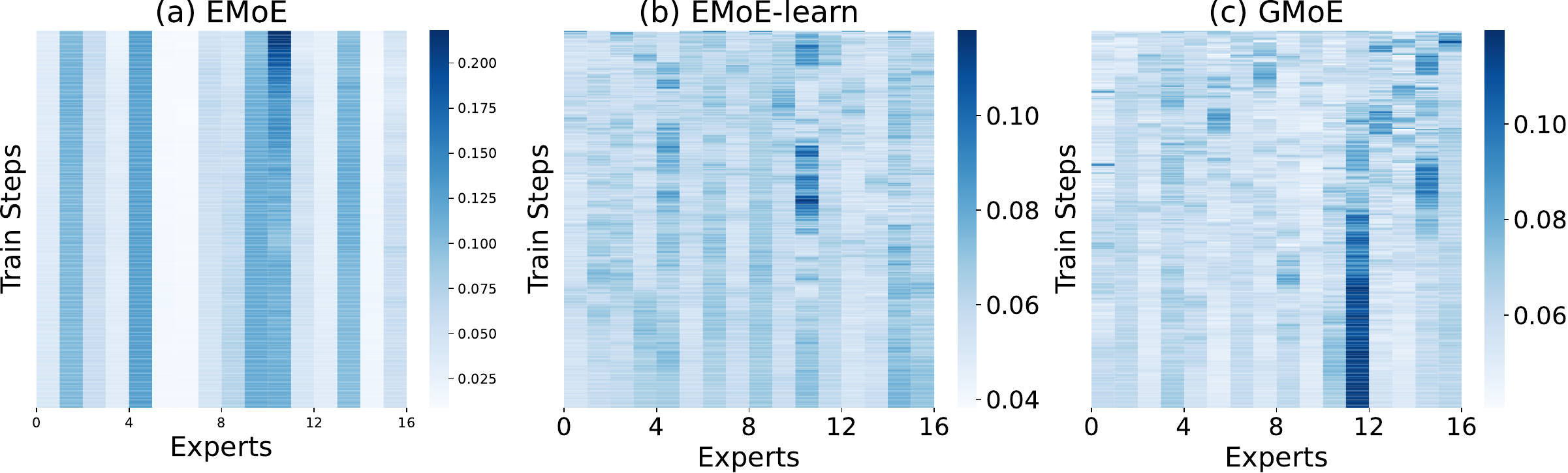}
}
\caption{Expert selections during training with distinct gating functions ($\operatorname{avg-k}$ vs. learned) and expert types (splits of FFNs vs. copies of FFNs).
The vertical axis illustrates training steps (top-down arrangement signifies begin-end); the horizontal axis represents expert selection frequency within 1K steps (deeper color implies a higher frequency).
(a), (b) and (c) correspond to EMoE, EMoE-learn, and GMoE.}
\label{fig:compare-gates-qnli}
\end{center}
\vskip -0.3in
\end{figure*}

Inspired by this, we choose a multi-task learning setting where negative transfer might be more pronounced. 
We adopt the codebase\footnote{https://github.com/AkariAsai/ATTEMPT} from ATTEMPT~\citep{DBLP:conf/emnlp/AsaiSPH22}.
For the ID scenario, we follow ATTEMPT and select six small tasks. 
For the OOD scenario, we train the models on two larger natural language inference (NLI) datasets and test them on four NLI datasets from different domains.
All hyperparameters unrelated to MoEs are consistent with the baseline.
We list the MoEs-related hyperparameters and average results in Tab.~\ref{tab:T5-results}.
More details about experiment setting and results are in Appendix~\ref{app:multi-tasks}.
We can observe that EMoE exhibits a more substantial improvement compared to the baseline. In the ID setting, the highest improvement reaches \textbf{7.56}, even considering the average performance across the six tasks. 
In the OOD setting, the highest average OOD result across the four datasets also improves by 1.58.

\begin{table}[h]
  \centering
  \caption{Results for different expert constructions (clustering-based v.s. random) and selections}
  \vspace{-0.65em}
  \label{tab:random-cluster}
  \resizebox{0.6\columnwidth}{!}{
  \begin{tabular}{c|cc}
    \toprule
    & Top-k &  Bottom-k\\
    \midrule
    Cluster & +0.92 &  -0.48 \\
    Random & -0.11 & -0.34 \\
    \bottomrule
  \end{tabular}
  }
  \vspace{-1.0em}
\end{table}

\begin{figure*}[t!]
\vskip -0.5in
\begin{center}
\centerline{\includegraphics[width=1.7\columnwidth]{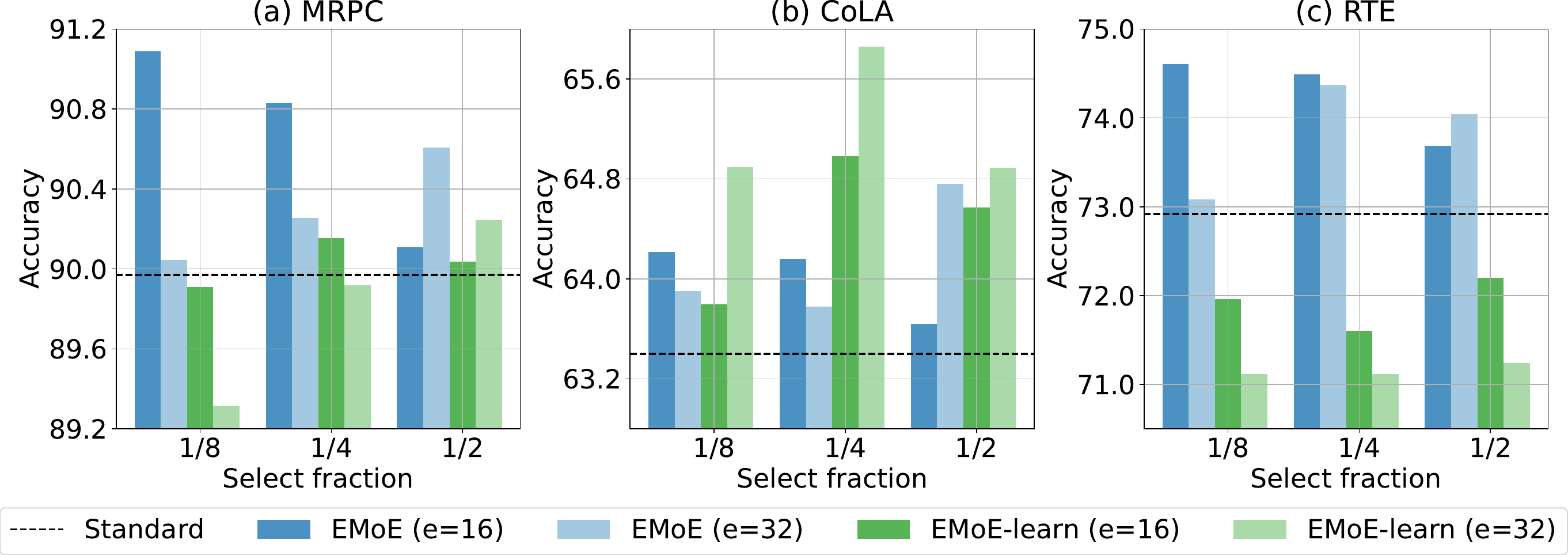}}
\caption{Results of 3 tasks for different experts splittings and expert select fractions (top-k / number of experts). `e=16' and `e=32' mean splitting FFNs into 16 and 32 experts, respectively.}
\label{fig:different_gates_configs}
\end{center}
\vskip -0.3in
\end{figure*}

\section{Ablation studies}
\label{sec:Analysis_Ablation}
\textbf{Sparsity and Modularity}
To provide further evidence that the EMoE’s improvements stem from leveraging modular features rather than just sparse activation or MoEs architecture, we compare the results of (1) \textit{key-vector-based clustering} expert construction and (2) \textit{random} construction. We employ the same setting of Sec.~\ref{para:Analysis_How_train}. 
The relative changes in averaged outcomes compared to the vanilla fine-tuning are shown in Tab.~\ref{tab:random-cluster}, while full results can be found in the appendix~\ref{appdendix:full_tables}, Tab.~\ref{tab:full_results_selections}. 
It's noteworthy that while cluster top-k exhibits a significant improvement over the standard, random top-k is conversely worse than vanilla fine-tuning.
This suggests that random construction can negatively impact gating, and only MoEs structure itself can't bring improvement.
Moreover, when selecting weights with negative transfer under bottom-k selections, it's observed that cluster bottom-k also achieves lower results.
In summary, \textbf{clustering-based methods can externalize implicit modularity within the pre-trained language model}.
Within suitable frameworks like MoEs, such modularity can facilitate downstream tuning.

\textbf{MoEs Constructing Methods} 
In EMoEs, the expert construction methods and gating methods are pivotal. 
To further understand splitting versus replicating FFNs, $\operatorname{avg-k}$ gating versus learned gating, we visualize expert selections of modularized GPT2-XL during tuning on 6 tasks with 16 experts.
In Fig.~\ref{fig:compare-gates-qnli}, we showcase the results for the largest dataset QNLI among them. Full results are available in the Appendix~\ref{appdendix:more_visualization}.
Our observations are:
(1) Both $\operatorname{avg-k}$ gating and learned gating converge, as indicated by the lower halves of the plots.
(2) $\operatorname{avg-k}$ gating is more stable than learned gating. As shown in Fig.~\ref{fig:compare-gates-qnli} (a) and (b), the expert selection merely changes during fine-tuning.
This might mitigate data inefficiency from gating inconsistencies across different training stages~\citep{DBLP:conf/iclr/Zuo00KHZGZ22}.
(3) EMoE, with its differently initialized experts parameters, exhibits better load balancing than GMoE (expert 11 in Fig.~\ref{fig:compare-gates-qnli} c is way more frequently selected than other experts). 
In GMoE, all experts share identical initialization; in EMoE, the experts are derived from FFNs with EM.
This also suggests a good initialization can facilitate MoEs learning, as indicated by~\citet{nie2021evomoe}.

Fig.~\ref{fig:compare-gates-qnli} (a) and Appendix~\ref{appdendix:more_visualization} demonstrate that some experts in EMoE are barely selected during fine-tuning.
This bears similarity to pruning, leading to the argument that EMoE's improvements may stem from pruning. 
In this context, continuing our exploration from \ref{para:Analysis_when} on examining EMoE during training and inference, we supplement two ablations under the same setting: 
1. \textit{Training-Pruning}: Training EMoE (N=64, top-k=32), pruning experts with lower selection frequency, and evaluating the pruned model.
2. \textit{Pruning-Training}: pruning lower selection frequency experts of a new model as per step 1, then training and evaluating the pruned model. 
The experimental outcomes are detailed in Tab.~\ref{tab:pruning_compare}. Our primary findings are:
1. When sparsely selected experts are pruned post-EMoE tuning, the performance consistently outperforms the LoRA baseline (even when only 1/16 experts are used). This underscores that EMoE's impact is manifested during the fine-tuning stage, not inference.
2. In the optimal pruning-training setting, the performance surpasses the baseline but falls significantly short of EMoE.
This suggests that EMoE's improvements are not solely attributable to pruning.
Compared with EMoE selecting experts for each token, pruning masks the same neurons for all tokens. 
Therefore, pruning can also be considered task-level EMoE, thus less effective.
Lastly, it is important to emphasize that in multi-task learning, pruning methods struggle to ascertain the neurons requiring pruning, whereas EMoE demonstrates significant effectiveness in these scenarios.

\textbf{MoEs Configurations} 
\label{para:Analysis_specialization}
Beyond the settings detailed in the main results section, which are based on $N=64$ experts and top-k $\in \{16, 32, 48\}$, we also present specific scenarios where $N \in \{16, 32\}$, and top-k varies within $\{2, 4, 8, 16\}$ in Fig.~\ref{fig:different_gates_configs}. 
Notably, in each of these settings, 
(1) EMoE consistently surpasses the standard model, illustrating its robustness to hyper-parameters.
(2) On average, $\operatorname{avg-k}$ gating exhibits superior performance than learned gating. Though learned gating (EMoE-learn) outperforms $\operatorname{avg-k}$ gating in a few specific settings (Fig.~\ref{fig:different_gates_configs} (b) and (e)).
This is consistent with the earlier results in Section~\ref{sec:main_results}.
Regarding how many EMoE layers should be introduced,
our findings align with those discovered in GMoE, indicating that only a limited number of layers can be converted into the EMoE layer. 
If excessive EMoE layers are introduced, performance deteriorates.
Taking GPT2-XL (48 layers) as an example, when introducing EMoE every two layers in the latter half, the performance averaged across 5 GLUE tasks (79.36) matches that of the standard model (78.87).
However, when adopting EMoE for every 2-layer for the entire model, the performance lags slightly behind that of the standard model (78.17) but surpasses the EMoE-learn (75.87). For additional configurations, please refer to the appendix~\ref{appdendix:full_analysis_results} Tab.~\ref{tab:results_appendix_more_MoEs_config}.

\section{Related Work}

Works most related to our work is introducing modularity based on off-the-shelf pre-trained models. 
For example, GMoE~\citep{DBLP:conf/iclr/LiSYWRCZ023} and Upcycling~\citep{DBLP:conf/iclr/KomatsuzakiPLRM23} copy the FFNs from a trained transformer model to form the MoEs architecture. 
Their modular structure is introduced by replicating existing FFNs modules, leaving EM within pre-trained FFNs unexplored. 
MoEfication~\citep{DBLP:conf/acl/ZhangL00S022} and MoEBert~\citep{zuo2022moebert} explores the EM within the model. 
They seek to improve the inference efficiency by decomposing the original FFNs layer into sparse MoEs.
They utilize the sparse expert activation to reduce inference overhead and do not touch on how EM influences the model's performance in downstream fine-tuning.
Our method to split FFNs is adopted from one simple method in the ablation studies of MoEfication paper~\citep{DBLP:conf/emnlp/ZhangSHZR022}: clustering-based expert construction and $\operatorname{avg-k}$ gating. 
We empirically find that such a simple method can validate the improvements brought by EM and answer our research questions. 
We thus leave more elaborated methods for future works. Please refer to Appendix~\ref{app:compare_MoEfication} for a detailed comparison of EMoE with those related works.

\section{Conclusion 
}

In this work, we validate that unlocking the EM in standard LMs improves downstream task ID and OOD performances.
EMoE can bring this benefit without adding any parameters, significant training costs, or any alterations to its deployment, which improves its practicability in the LLMs era.
One possible reason is the modular structure can alleviate negative transfer effects presented in the LMs. 
We hope our findings can deepen the understanding of neural networks' modularity, further helping the community develop more sophisticated modular neural architectures and utilize existing LMs.

\section{Limitations}

Our primary objective was to investigate the utility of EM, and thus, we predominantly adopted the techniques from MoEfication for decomposition. 
We encourage further research to propose improved algorithms for harnessing EM.
Our research findings have not been validated on more challenging tasks (e.g., Mathematical Reasoning~\citep{DBLP:conf/acl/ImaniD023}).
While our analysis was primarily conducted on models with a maximum parameter count of 1.5B, we validate the scaling-up ability of EMoE to Llama-30B.

\bibliography{custom}

\appendix

\section{Additional Information}

\subsubsection{Constrained clustering}
\label{appendix:constrained}
Constrained clustering~\citep{DBLP:conf/sspr/MalinenF14} enhances traditional clustering methods by incorporating additional user-specified constraints. This approach involves balancing the k-means algorithm for clustering, making it more efficient and effective in generating meaningful clusters. The introduction of constraints, such as must-link or cannot-link conditions, guides the clustering algorithm to produce results that align more closely with domain-specific requirements or prior knowledge. This method has improved the relevancy and accuracy of clustering outcomes in various applications.

\section{Additional Experiment Results}

\subsection{Full Fine-tuning Performance}
\label{appendix:full-ft}

\begin{table*}[htbp!]
\small
    \setlength{\tabcolsep}{4pt}
    \renewcommand{\arraystretch}{1.0}
    \centering
    \caption{Overall OOD performances with 3 selection criteria.
    All the reported results are obtained following the Domainbed codebase. 
    The best result is highlighted in \textbf{bold}.
    In cases where results are the same, the best result is determined by the smallest standard deviation. 
    EMoE demonstrates comparable results to GMoE.}
    \label{tab:main_results_domainbed}
    \begin{minipage}[t]{0.43\textwidth}
      \centering
      \captionsetup{font=small}
      \caption*{Results with ViT-small (22M) backbone}
      \resizebox{\textwidth}{!}
      {%
      \begin{tabular}{l|cccccc}
      \toprule
      \rowcolor{COLOR_MEAN}
      \textbf{Algorithm} & \textbf{PACS} & \textbf{VLCS} & $\scalebox{0.8}{\textbf{Office}}$ & $\scalebox{0.9}{\textbf{Terra}}$ &\textbf{Avg} \\ \midrule
      \multicolumn{6}{c}{Train-validation selection criterion} \\ \midrule
      ViT & 86.9 & \textbf{79.7} & 73.0 & 44.0 & 70.90\\
      GMoE & 87.7 & 79.6 & 73.1 & 45.4 & 71.45\\
     $\scalebox{0.9}{EMoE-learn}$ & 87.2 & 79.6 & 72.5 &  \textbf{46.1} &71.35\\
     \rowcolor{ROW_COLOR}
      EMoE & \textbf{87.8} & 79.5 &  \textbf{73.1} & 45.9 &\textbf{71.58}\\ \midrule
      \multicolumn{6}{c}{Leave-one-domain-out selection criterion} \\ \midrule
      ViT & 86.1 & 79.7 & \textbf{73.3} & 45.0 &71.03\\
      GMoE & 86.5 & 80.5 & 73.1 & 45.3 &71.35\\
     $\scalebox{0.9}{EMoE-learn}$ &  \textbf{86.8} & 79.6 & 72.6 & 45.8 &71.20\\
     \rowcolor{ROW_COLOR}
      EMoE & 86.8 & \textbf{80.6} & 73.3 &  \textbf{46.1} &\textbf{71.70}\\ \midrule
      \multicolumn{6}{c}{Test-domain selection criterion} \\ \midrule
      ViT & 86.5 & 78.2 & 73.1 & 44.0 &70.45\\
      GMoE & 87.2 & 79.0 &  \textbf{73.4} & 45.3 &71.23\\
     $\scalebox{0.9}{EMoE-learn}$ & 87.4 &  \textbf{79.1} & 72.8 & 45.4 &71.18\\
     \rowcolor{ROW_COLOR}
      EMoE & \textbf{87.6} & 79.0 & 73.3 &  \textbf{45.5} & \textbf{71.35}\\
      \bottomrule
      \end{tabular}%
      }
    \end{minipage}%
    \hspace{0.5em}
    \begin{minipage}[t]{0.43\textwidth}
      \centering
      \captionsetup{font=small}
      \caption*{Results with ViT-base (86M) backbone}
      \resizebox{\textwidth}{!}{%
      \begin{tabular}{l|cccccc}
      \toprule
      \rowcolor{COLOR_MEAN}
      \textbf{Algorithm} & \textbf{PACS} & \textbf{VLCS} & $\scalebox{0.8}{\textbf{Office}}$ & $\scalebox{0.9}{\textbf{Terra}}$ &\textbf{Avg} \\ \midrule
      \multicolumn{6}{c}{Train-validation selection criterion} \\ \midrule
      ViT & 89.1 &  \textbf{80.7} & 77.2 & 47.3 & 73.58 \\
      GMoE &  \textbf{90.0} & 80.4 & 77.0 & \textbf{49.2} & \textbf{74.15} \\
     $\scalebox{0.9}{EMoE-learn}$ & 89.8 & 80.6 & 76.5 & 48.7 & 73.90 \\
     \rowcolor{ROW_COLOR}
      EMoE & 89.4 & 80.7 & \textbf{77.3} & 48.5 &73.98 \\ \midrule
      \multicolumn{6}{c}{Leave-one-domain-out selection criterion} \\ \midrule
      ViT & 88.9 & 80.8 & \textbf{77.5} & 46.1 &73.33 \\
      GMoE & 89.3 & 81.0 & 76.7 & 50.1  &74.28\\
     $\scalebox{0.9}{EMoE-learn}$ & 89.3 & 81.2 & 76.5 & \textbf{50.5} &74.38 \\
     \rowcolor{ROW_COLOR}
      EMoE & \textbf{89.6} & \textbf{81.6} & 77.4 & 50.0  &\textbf{74.65}\\ \midrule
      \multicolumn{6}{c}{Test-domain selection criterion} \\ \midrule
      ViT & 88.8 & 79.0 & 77.2 & 46.7 &72.93\\
      GMoE & 89.7 & 79.0 & 77.0 & \textbf{48.8} &73.63\\
     $\scalebox{0.9}{EMoE-learn}$ & \textbf{89.7} & \textbf{79.7} & 76.6 & 48.7 &73.68\\
     \rowcolor{ROW_COLOR}
      EMoE & \textbf{89.7} & \textbf{79.7} & \textbf{77.5} & 48.8 &\textbf{73.93}\\
      \bottomrule
      \end{tabular}%
      }
    \end{minipage}
\end{table*}

\textbf{We test the EMoE's OOD performance on Domainbed.} 
\label{para:Domainbed_config}
Domainbed provides comprehensive vision OOD evaluations (one result is aggregated with 30 experiments), and the outcomes vary marginally. 
Moreover, ~\citet{DBLP:conf/iclr/GulrajaniL21} indicates that vanilla full fine-tuning with fair hyper-parameter search is a strong baseline compared with specifically designed methods like Invariant Risk Minimization~\citep{DBLP:journals/corr/abs-1907-02893}. 
More dataset details are in appendix~\ref{appendix:Domainbed}.
According to Table~\ref{tab:main_results_domainbed}: (1) Overall, EMoE outperforms ViT and GMoE (except ViT-base Train-validation setting, upper right).
(2) Compared with EMoE, EMoE-learn incorporates a learned gate. While it surpasses $\operatorname{avg-k}$ gating in certain scenarios (ViT-small Terra Train-validation), it can also lead to a performance drop compared with vanilla ViT. Its overall performance is lower than EMoE.
(3) In tasks where the standard model is strong (like Office), EMoE performs better than other MoEs methods.
One possible reason is that the $\operatorname{avg-k}$ gating reduces the influence of gating weights ($g_n(\mathbf{x};\mathbf{G};k)$ in Equation~\ref{eqn:MoE}), making it more like the standard model in such scenarios.

In vision tasks, as highlighted by~\citep{DBLP:conf/iclr/GulrajaniL21}, the hyper-parameter search has a profound impact on outcomes. 
Consequently, we search with a relatively large scope: for GMoE, $N$ is searched within \{4, 6, 8\}, and top-k within \{2, 3, 4\}; for EMoE, $N$ is sought within \{6, 12, 24\}, and top-k within \{2, 4, 8\}.

\begin{table*}[thbp!]
\small
    \setlength{\tabcolsep}{4pt}
    \renewcommand{\arraystretch}{1.0}
    \centering
    \caption{ID performance on GLUE tasks with different backbones and algorithms. All the reported results are obtained from 3 independent experiments. The average accuracy (Avg) is reported along with the relative improvement compared to the baseline. The best result is highlighted in \textbf{bold}.}
    \label{tab:results_full_tuning_language}
    \resizebox{0.7\textwidth}{!}{
    \begin{tabular}{c|l|cccccc}
    \toprule
    \rowcolor{COLOR_MEAN}
    \textbf{$\scalebox{0.8}{Backbone}$} & \textbf{$\scalebox{0.8}{Algorithm}$} & \textbf{MRPC} & \textbf{CoLA} & \textbf{RTE} & \textbf{STSB} & \textbf{SST2} & \textbf{Avg} \\
    \midrule
    \multirow{5}{*}{\parbox{1cm}{\centering $\scalebox{0.9}{BERT}$ \\ $\scalebox{0.9}{Base}$}} & baseline & 88.45 & 60.67 & 68.95 & 87.87 & 91.97 & 79.582 \\
    &$\scalebox{0.9}{noisy tuning}$ & 88.43 & 61.79 & 71.36 & 88.27 & 92.32 & 80.43(+0.85)\\
    & GMoE & 88.63 & 61.25 & 70.28 & 88.63 & 92.28 & 80.21(+0.63)\\
    &$\scalebox{0.9}{EMoE-learn}$ & 89.05 & \textbf{62.46} & \textbf{70.40} & 88.47 & 92.58 & \textbf{80.59(+1.01)} \\
     \rowcolor{ROW_COLOR}
    & EMoE & \textbf{89.45} & 61.55 & 69.68 & \textbf{88.71} & \textbf{92.89} & 80.46(+0.87)\\
    \midrule
    \multirow{5}{*}{\parbox{1cm}{\centering $\scalebox{0.9}{BERT}$ \\ $\scalebox{0.9}{Large}$}} & baseline & 89.82 & 65.41 & 74.89 & 89.87 & 93.50 & 82.70 \\
    &$\scalebox{0.9}{noisy tuning}$ & 90.42 & 64.75 & 73.41 & 90.05 & 93.65 & 82.46(-0.24)\\
    & GMoE & \textbf{91.24} & 64.90 & 74.24 & 90.00 & 93.58 & 82.79(+0.09)\\
    &$\scalebox{0.9}{EMoE-learn}$ & 90.57 & 65.51 & 74.72 & 90.22 & \textbf{93.73} & 82.95(+0.25)\\
     \rowcolor{ROW_COLOR}
    & EMoE & 90.74 & \textbf{65.79} & \textbf{76.17} & \textbf{90.31} & 93.58 & \textbf{83.32(+0.62)}\\
    \midrule
    \multirow{5}{*}{\parbox{1cm}{\centering $\scalebox{0.9}{GPT2}$ \\ $\scalebox{0.9}{Small}$}} & baseline & 84.46 & 47.07 & 67.15 & 86.29 & 92.13 & 75.42 \\
    &$\scalebox{0.9}{noisy tuning}$ & 84.15 & 46.16 & 67.51 & 86.09 & 92.13 & 75.21(-0.21)\\
    & GMoE & 85.07 & 47.77 & 67.51 & 86.57 & 92.35 & 75.85(+0.43)\\
    &$\scalebox{0.8}{EMoE-learn}$ & \textbf{85.73} & 47.24 & 67.99 & \textbf{86.66} & 92.35 & 75.99(+0.57)\\
     \rowcolor{ROW_COLOR}
    & EMoE & 85.40 & \textbf{48.00} & \textbf{68.95} & 86.64 & \textbf{92.70} & \textbf{76.34(+0.92)}\\
    \bottomrule
    \end{tabular}
    }
\end{table*}

\textbf{We evaluate EMoE's ID performance on 5 GLUE tasks}. 
We also include an additional baseline noisy tuning~\citep{DBLP:conf/acl/WuWQ022}, which improves adaptation for free by adding uniform distribution noise proportional to the standard deviation of the pre-trained weights before tuning.
\label{para:Full_configs}
According to Table~\ref{tab:results_full_tuning_language}:
(1) On average, EMoE and EMoE-learn outperform other baselines.
(2) Among the two methods that do not introduce additional parameters, EMoE significantly outperforms noisy tuning.
(3) EMoE provides stable improvements over baselines across different settings, demonstrating its generality.

\begin{table}[h]
\centering
\caption{Instruction full-tuned Llama2-7B's MMLU scores. Times are wall-clock computation times. The term 'times' in the subsequent tables refers to the same concept. 'w/o' refers to 'without'.}
\label{tab:llama2-7B-ft}
\resizebox{0.5\textwidth}{!}{
\begin{tabular}{c|c|c|c|c}
\toprule
 Experts (N) & topk (k) & MMLU score & times (s) & FLOPS ($10^{16}$) \\
\midrule
w/o tuning & - & 46.79 & - & - \\
full tuning & - & 46.5 & 4988 & 8.97 \\
64 & 16 & \textbf{48.08} & 5036 & 9.12 \\
64 & 32 & 47.44 & 5041 & 9.24 \\
\bottomrule
\end{tabular}
}
\end{table}

\subsection{Mult-tasks Setting}
\label{app:multi-tasks}

\begin{table}[ht]
\centering
\caption{T5-base ID performances. All tasks are from SuperGLUE so we omit the prefix "SuperGLUE" for each tasks.}
\resizebox{0.5\textwidth}{!}{\begin{tabular}{c|c|c|c|c|c|c|c|c}
\toprule
Experts (N) & topk (k) & boolq & cb & wic & wsc.fixed & rte & copa & test avg \\
\midrule
Baseline & & 82.14 & 85.71 & 65.83 & 34.62 & 74.10 & 52.00 & 65.73 \\
8 & 2 & 80.67 & 89.29 & 65.52 & 36.54 & 79.86 & 56.00 & 67.98 \\
16 & 4 & 81.16 & 89.29 & 68.34 & 51.92 & 74.10 & 44.00 & 68.14 \\
32 & 8 & 80.12 & 78.57 & 70.85 & 63.46 & 82.73 & 64.00 & \textbf{73.29} \\
32 & 16 & 81.04 & 75.00 & 72.41 & 57.69 & 78.42 & 54.00 & 69.76 \\
\bottomrule
\end{tabular}
}
\end{table}

\begin{table}[htbp]
    \centering
    \caption{T5-base OOD performances. 'SG' refers to 'SuperGLUE'.}
    \label{tab:t5-ood-performances}
    \resizebox{0.5\textwidth}{!}{
    \begin{tabular}{c|c|c|c|c|c|c|c|c}
        \toprule
        Experts (N) & topk (k) & mnli & qnli & wnli & rte & SG-rte & SG-cb & OOD avg \\
        \midrule
        Baseline & & 86.2 & 92.42 & 50.00 & 61.87 & 62.59 & 32.14 & 51.65 \\
        8 & 2 & 86.27 & 92.18 & 50.00 & 64.03 & 62.59 & 32.14 & 52.19 \\
        16 & 2 & 86.44 & 92.59 & 52.78 & 64.03 & 56.83 & 39.29 & \textbf{53.23} \\
        32 & 8 & 86.56 & 92.49 & 58.33 & 64.03 & 61.87 & 28.57 & 53.20 \\
        \bottomrule
    \end{tabular}
    }
\end{table}

In our analysis~\ref{sec:analysis}, we have identified that the improvement brought by EMoE is likely associated with mitigating negative transfer. 
Inspired by this, we choose a multi-task learning setting where negative transfer might be more pronounced. 
We adopt the codebase\footnote{https://github.com/AkariAsai/ATTEMPT} from ATTEMPT~\citep{DBLP:conf/emnlp/AsaiSPH22}.
For the in-domain (ID) scenario, we follow the settings outlined in ATTEMPT and select six tasks from the Super-GLUE benchmark~\citep{DBLP:conf/nips/WangPNSMHLB19}. 
For the out-of-domain (OOD) scenario, we take two larger natural language inference (NLI) datasets MNLI and QNLI, from GLUE as our ID training data. We subsequently conducted direct testing on four additional NLI datasets from different domains. 
All hyperparameters unrelated to MoEs are kept consistent with the baseline, and we have listed the MoEs-related hyperparameters in the result table.
Our observations are as follows:
1. EMoE exhibits a substantial improvement compared to the baseline. In the in-domain (ID) setting, the highest improvement reached 7.56, even considering the average performance across the six tasks. In the OOD setting, the highest average OOD result across the four datasets improved by 1.58.
2. Across various settings of N and K, EMoE consistently outperforms the vanilla fone-tuning. Within the hyperparameter search space specified in our paper, EMoE consistently improves at least 2 points over the baseline in the ID setting. This emphasizes the effectiveness of EMoE and EMoE’s robustness to the explored hyperparameter range.

\subsection{Performance Across Different Training set Volumes}
\begin{figure}
  \centering
  \includegraphics[width=0.4\textwidth]{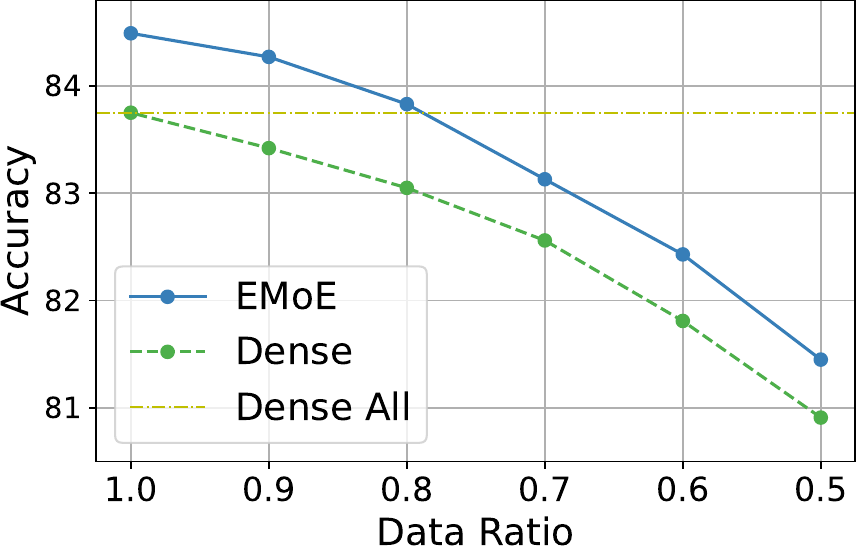}
  \caption{Average performance of EMoE with different proportions of training data.}
  \label{fig:different_data}
\end{figure}

Previous research has indicated that modular architectures offer improved data efficiency~\citep{DBLP:conf/iclr/BengioDRKLBGP20}.
Therefore, we conducted experiments with GPT2-XL on six tasks using varying proportions of original training data, and the results for all tasks are presented in Figure~\ref{fig:different_data}. 
It can be observed that EMoE consistently outperforms the standard across different data factions.
EMoE achieves superior results even when using less than 20\% of the data. On SST2, only using 50\% data, EMoE shows comparable performance to the standard. More details can be found in the Appendix~\ref{appdendix:full_analysis_results} Table~\ref{tab:results_comparison_datafraction}.
This further underscores the benefits of incorporating modular structures.

\section{Comparison between EMoE, MoEfication, and GMoE}
\label{app:compare_MoEfication}

\begin{table*}[h]
  \centering
  \caption{Comparison between EMoE, MoEfication, and GMoE}
  \label{tab:emoe_vs_moefication}
  \resizebox{0.9\textwidth}{!}{
  \begin{tabular}{p{3cm}|p{4cm}|p{4cm}|p{4cm}}
    \toprule
    \small
    \textbf{Aspect} & \textbf{EMoE} & \textbf{MoEfication} &\textbf{GMoE} \\
    \midrule
    Research Problem & Exploit the emergent modularity during fine-tuning pre-trained transformers. & Approximate FFNs with sparse MoE to improve inference efficiency. & Validate the OOD improvements brought by Sparse MoE architectures. \\
    \midrule
    Methods & Split FFNs & Split FFNs & Copy of FFNs\\
    \midrule
    Practicality & No additional trainable parameters. Experts can be recomposed into the standard model so that models can be deployed as a standard model. & May need re-training on the original task.  May suffer from inference latency owing to the specific implementation of MoE architectures. & Additional trainable parameters are introduced. May suffer from inference latency owing to the specific implementation of MoE architecture. \\
    \midrule
    Contribution& Significant general improvement without adding parameters and not depending on the specific implementation. & Improved inference efficiency (depending on the specific implementation of MoE), but performance drop. & Significant OOD improvement with additional parameters and specific implementation. \\
    \bottomrule
  \end{tabular}
  }
\end{table*}

Leveraging modular designing into neural networks has various advantages~\citep{DBLP:journals/corr/abs-2302-11529,DBLP:journals/corr/abs-2204-02311,DBLP:journals/corr/abs-2306-17165}, including superior generalization abilities~\citep{DBLP:conf/iclr/GoyalLHSLBS21, DBLP:conf/iclr/LiSYWRCZ023}.
Most MNNs are \textit{explicitly} modular. Among them, MoEs~\citep{DBLP:conf/icnn/SzymanskiL93} is currently a standard architecture for developing MNNs~\citep{DBLP:journals/corr/abs-2306-04640,DBLP:conf/iclr/ShazeerMMDLHD17,DBLP:journals/jmlr/FedusZS22,DBLP:conf/emnlp/ZhangSHZR022}.
Apart from explicit MNNs, ~\citet{DBLP:journals/corr/abs-2110-08058, DBLP:conf/iclr/CsordasSS21} explore emergent modular structures in CNNs and LSTMs. 
Some recent works~\citep{DBLP:conf/acl/ZhangL00S022,DBLP:journals/corr/abs-2210-06313} focus on the sparsity of more complicated pre-trained transformers. 
Based on their observations, ~\citet{DBLP:conf/acl/ZhangZLXW00XS023} recently explore modularity in pre-trained transformer FFNs by employing handpicked semantic and knowledge-intensive tasks to probe the modular nature of pre-trained transformers.

Having observed that FFN layers in pre-trained Transformers are sparsely activated (many neurons are unused for inputs), MoEfication splits transforms FFNs into a sparse MoE, aiming to approximate the functionality of the original FFNs to reduce the computational cost, further improving inference efficiency. 
Besides, the GMoE makes multiple replicates of the original FFN layer and introduces a learned gate to form a MoE architecture. They claim that such architecture could improve OOD performance from their theoretical perspective of algorithmic alignment framework. 
MoEfication and GMoE do not touch on how emergent modularity influences the training stage. The table below illustrates a detailed comparison of these works.
These differences are summarized in Table~\ref{tab:emoe_vs_moefication}.

\section{Datasets and Evaluation Metrics}
\label{appendix:datasets_eval}
\subsubsection{Domainbed}
\label{appendix:Domainbed}

\begin{table}[h]
  \centering
  \caption{Used dataset information from Domainbed}
  
  \label{tab:dataset_stats}
  \resizebox{0.5\textwidth}{!}
    {
  \begin{tabular}{lcccc}
    \toprule
    \textbf{Dataset} & \textbf{PACS} & \textbf{VLCS} & \textbf{OfficeHome} & \textbf{TerraInc} \\
    \midrule
    \#Domains & 4 & 4 & 4 & 4 \\
    Classes & 7 & 5 & 65 & 10 \\
    Images & 9,991 & 10,729 & 15,588 & 24,788 \\
    \bottomrule
  \end{tabular}
  }
\end{table}

The four datasets (PACS~\citep{DBLP:journals/corr/abs-1710-03077}, VLCS~\citep{DBLP:conf/iccv/FangXR13}, Office-Home~\citep{DBLP:conf/cvpr/VenkateswaraECP17}, and Terra Incognita~\citep{DBLP:conf/eccv/BeeryHP18}) are selected from Domainbed. 
Each dataset comprises 4 distinct domains(PACS: \{art, cartoons, photos, sketches\}, VLCS: \{Caltech101, LabelMe, SUN09, VOC2007\}, Office-Home: \{art,clipart, product, real\}, and Terra Incognita: \{L100, L38, L48, L46\}).
One or two domains' data are sequentially designated within a single training for OOD evaluation. 
For example, when training on PACS, \{art, cartoons\} could be selected as ID training data, while \{photos, sketches\} are designated for OOD testing. 
This configuration results in $C_4^2+C_4^1=10$ distinct training processes within each dataset. 
Suppose there are $d_{tr}$ ID domains, ``Train-validation'' means selecting OOD test checkpoints based on ID accuracies from the validation subsets of all $d_{tr}$ ID domains; 
``Leave-one-domain-out'' means leaving one selected ID domain as a validation set, doing training on $d_{tr}-1$ domains; 
``Test-domain'' means selection based on limited access to test domains and selection based on these results. 
The final results are aggregated with selection criteria provided by Domainbed\footnote{https://github.com/facebookresearch/DomainBed}. 
As a result, even a variation of 0.1 in the benchmark outcomes signifies a significant improvement.

In our experiments, all the hyper-parameters, like training steps, learning rates, and weight decay, except those related to MoEs, strictly follow GMoE.

\subsection{GLUE}
\label{appendix:GLUE}

Each task involves one to four OOD tasks from GLUE-X~\citep{DBLP:conf/acl/YangZ0LWLW0023}, resulting in 13 OOD tasks in total.
To illustrate, consider the Sentiment Analysis task: we first fine-tune models on SST-2 from GLUE and report the validation results as ID performance, then use the test data from IMDB~\citep{DBLP:conf/acl/MaasDPHNP11}, Yelp~\citep{DBLP:conf/nips/ZhangZL15}, Amazon~\citep{DBLP:conf/iclr/KaushikHL20} and Flipkart~\citep{thummar_2023} from GLUE-X for OOD testing.

In the \textit{full fine-tuning}, to ensure convergence and reduce randomness, we train all models 10 epochs across 3 random seeds on each task. 
Each experiment does a hyper-parameter search on learning rates on [2e-5, 3e-5, 5e-5] as suggested by BERT~\citep{DBLP:conf/naacl/DevlinCLT19}. 
The training batch size is 32. 
In the \textit{LoRA tuning}, following LoRA~\citep{DBLP:conf/iclr/HuSWALWWC22} that tunes models with more epochs and larger learning rates than standard full fine-tuning, all models are trained 20 epochs on small and medium datasets and 5 epochs on large ones (like QNLI, MNLI, QQP). 
The learning rate is searched in [2e-4, 3e-4, 5e-4].  All methods are implemented with LoRA\_rank=8 and LoRA\_alpha=16.
The training batch size is 16 due to a larger model size. 
Other settings like max\_lengt following the codebase from hugging face\footnote{https://github.com/huggingface/transformers}.
After training on GLUE, we directly test the selected models on GLUE-X with the data from the official repo\footnote{https://github.com/YangLinyi/GLUE-X}.

\begin{table*}[!ht]
    \centering
    \caption{Language tasks and corresponding ID and OOD datasets.}
    \resizebox{0.7\textwidth}{!}
    {
    \begin{tabular}{llclc}
    \toprule
        Task & ID-dataset & size & OOD-dataset & size \\ 
        \midrule
        \multirow{3}{*}{Paraphrase}  & MRPC  & 4,076 & Twitter  & 16,777 \\ 
        ~ & \multirow{2}{*}{QQP}  & \multirow{2}{*}{404,301} & Twitter  & 16,777 \\ 
        ~ & ~ & ~ & MRPC & 4,076 \\ 
        \midrule
        Linguistic Acceptability  & CoLA  & 9,594 & Grammar Test & 304,277 \\ 
        \midrule
        \multirow{2}{*}{Textual Entailment}  & \multirow{2}{*}{RTE}  & \multirow{2}{*}{2,768} & SciTail  & 26,527 \\ 
        ~ & ~ & ~ & HANs & 60,000 \\ 
        \midrule
        Textual Similarity  & STSB & 7,128 & SICK  & 9,840 \\ 
        \midrule
        \multirow{4}{*}{Sentiment Analysis} & \multirow{4}{*}{SST2} & \multirow{4}{*}{68,223} & IMDB & 50,000 \\ 
        ~ & ~ & ~ & Yelp & 598,000 \\ 
        ~ & ~ & ~ & Amazon & 4,000,000 \\ 
        ~ & ~ & ~ & Flipkart & 205,041 \\
        \midrule
        Question Answering NLI  & QNLI & 110,206 & NewsQA & 119,525 \\ 
        \midrule
        Natural Language Inference  & MNLI & 412,313 & SICK  & 9,840 \\
        \bottomrule
    \end{tabular}
    }
\end{table*}

\section{Computation Cost and Memory Usage}
\label{appendix:computation_and_memory}

\subsection{Experiments with public MoEs library}

Theoretically, EMoE does not introduce additional parameters compared to the standard model. 
Although it adds computation in the gating portion within the MoEs layer, it omits a substantial amount of computation within the FFNs layer. 
For instance, the computation in the gating portion is of the order of $h\times N$, where h represents the model's hidden size, and N is the number of experts. 
In contrast, the complete computation in the FFNs layer is of the order of $(h \times h \times 4h) \times 2$, and sparse activations can reduce more than a quarter of this computation. 
Since $N \ll h$, theoretically, using EMoE within a single block should accelerate the forward pass of the model.
However, in real deployment, we have observed that the hardware implementation of MoEs can result in EMoE being, on average, approximately 10\% slower than the standard model. The memory usage is also slightly higher, by less than 5\% 

Each experiment was conducted on a single NVIDIA 40G A100 GPU. 
The training times for different tasks ranged from just over ten minutes (RTE) to more than ten hours (QQP).

\subsection{Experiments with self-implementation}

We further find that the increasing wall time and the GPU memory usage come from the public library tutel we used to implement EMoE. 
We reimplemented our method and observed that EMoE does not require significant additional run time and memory usage. 
Specifically, we introduce an alternative implementation approach in EMoE where hidden states are used to calculate gate scores after computing the first layer. 
These scores mask the outputs of unselected experts, mimicking the effect of MoEs.
Though this theoretically increases FLOPS compared to traditional MoEs, in practice, the speed is comparable to standard models, as demonstrated in Tab.~\ref{tab:llama2-7B-ft},~\ref{tab:llama-lora}
All experiments about LLMs are conducted on eight NVIDIA 80G A100 GPU. 

\section{Tabular Results}

\subsection{Full Tables in Full Fine-tuning with Standard Deviation}
\label{appdendix:full_tables}
This section presents the mean and variance of experiments conducted with three different random seeds. The Domainbed results are demonstrated in Table~\ref{tab:full_main_results_domainbed_std}, full fine-tuning results are in Table~\ref{tab:results_full_tuning_language_with_std}, LoRA tuning results are in Table~\ref{tab:full_results_lora_ID_OOD_std}.

\begin{table*}[h]
  \setlength{\tabcolsep}{4pt}
  \renewcommand{\arraystretch}{1.0}
  \centering
  \caption{Overall out-of-domain performances with different selection criteria. All the reported results are obtained from three independent experiments following the Domainbed codebase. The best result is highlighted in \textbf{bold}.In cases where results are the same, the smallest standard deviation determines the best result. EMoE demonstrates comparable results to GMoE.}
  \vspace{-0.4em}
  \label{tab:full_main_results_domainbed_std}
  \begin{minipage}[t]{0.49\textwidth}
    \centering
    \captionsetup{font=small}
    \caption*{Results with ViT-small backbone}
    \vspace{-0.7em}
    \resizebox{\textwidth}{!}
    {%
    \begin{tabular}{l|cccccc}
    \toprule
    \textbf{Algorithm} & \textbf{PACS} & \textbf{VLCS} & \textbf{OfficeHome} & \textbf{TerraInc} &\textbf{Avg} \\ \midrule
    \multicolumn{6}{c}{train-validation selection criterion} \\ \midrule
    ViT & 86.9$\scalebox{0.8}{$\pm$0.2}$ & \textbf{79.7$\scalebox{0.8}{$\pm$0.4}$} & 73.0$\scalebox{0.8}{$\pm$0.2}$ & 44.0$\scalebox{0.8}{$\pm$1.1}$ & 70.90\\
    GMoE & 87.7$\scalebox{0.8}{$\pm$0.2}$ & 79.6$\scalebox{0.8}{$\pm$0.4}$ & 73.1$\scalebox{0.8}{$\pm$0.3}$ & 45.4$\scalebox{0.8}{$\pm$0.3}$ & 71.45\\
   $\scalebox{0.9}{EMoE-learn}$ & 87.2$\scalebox{0.8}{$\pm$0.4}$ & 79.6$\scalebox{0.8}{$\pm$0.2}$ & 72.5$\scalebox{0.8}{$\pm$0.2}$ &  \textbf{46.1$\scalebox{0.8}{$\pm$0.4}$} &71.35\\
    EMoE & \textbf{87.8$\scalebox{0.8}{$\pm$0.2}$} & 79.5$\scalebox{0.8}{$\pm$0.4}$ &  \textbf{73.1$\scalebox{0.8}{$\pm$0.2}$} & 45.9$\scalebox{0.8}{$\pm$0.3}$ &\textbf{71.58}\\ \midrule
    \multicolumn{6}{c}{leave-one-domain-out selection criterion} \\ \midrule
    ViT & 86.1$\scalebox{0.8}{$\pm$0.6}$ & 79.7$\scalebox{0.8}{$\pm$0.4}$ & \textbf{73.3$\scalebox{0.8}{$\pm$0.1}$} & 45.0$\scalebox{0.8}{$\pm$0.5}$ &71.03\\
    GMoE & 86.5$\scalebox{0.8}{$\pm$0.3}$ & 80.5$\scalebox{0.8}{$\pm$0.2}$ & 73.1$\scalebox{0.8}{$\pm$0.3}$ & 45.3$\scalebox{0.8}{$\pm$0.6}$ &71.35\\
   $\scalebox{0.9}{EMoE-learn}$ &  \textbf{86.8$\scalebox{0.8}{$\pm$0.0}$} & 79.6$\scalebox{0.8}{$\pm$0.3}$ & 72.6$\scalebox{0.8}{$\pm$0.2}$ & 45.8$\scalebox{0.8}{$\pm$0.6}$ &71.20\\
    EMoE & 86.8$\scalebox{0.8}{$\pm$0.1}$ & \textbf{80.6$\scalebox{0.8}{$\pm$0.4}$} & 73.3$\scalebox{0.8}{$\pm$0.2}$ &  \textbf{46.1$\scalebox{0.8}{$\pm$0.6}$} &\textbf{71.70}\\ \midrule
    \multicolumn{6}{c}{test-domain selection criterion} \\ \midrule
    ViT & 86.5$\scalebox{0.8}{$\pm$0.4}$ & 78.2$\scalebox{0.8}{$\pm$0.4}$ & \textbf{73.1$\scalebox{0.8}{$\pm$0.2}$} & 44.0$\scalebox{0.8}{$\pm$0.5}$ &70.45\\
    GMoE & 87.2$\scalebox{0.8}{$\pm$0.4}$ & 79.0$\scalebox{0.8}{$\pm$0.3}$ &  \textbf{73.4$\scalebox{0.8}{$\pm$0.2}$} & 45.3$\scalebox{0.8}{$\pm$0.4}$ &71.23\\
   $\scalebox{0.9}{EMoE-learn}$ & 87.4$\scalebox{0.8}{$\pm$0.2}$ &  \textbf{79.1$\scalebox{0.8}{$\pm$0.3}$} & 72.8$\scalebox{0.8}{$\pm$0.1}$ & 45.4$\scalebox{0.8}{$\pm$0.6}$ &71.18\\
    EMoE & \textbf{87.6$\scalebox{0.8}{$\pm$0.5}$} & 79.0$\scalebox{0.8}{$\pm$0.2}$ & 73.3$\scalebox{0.8}{$\pm$0.0}$ &  \textbf{45.5$\scalebox{0.8}{$\pm$0.1}$} & \textbf{71.35}\\
    \bottomrule
    \end{tabular}%
    }
  \end{minipage}%
  \hspace{0.5em}
  \begin{minipage}[t]{0.49\textwidth}
    \centering
    \captionsetup{font=small}
    \caption*{Results with ViT-base backbone}
    \vspace{-0.7em}
    \resizebox{\textwidth}{!}{%
    \begin{tabular}{l|cccccc}
    \toprule
    \textbf{Algorithm} & \textbf{PACS} & \textbf{VLCS} & \textbf{OfficeHome} & \textbf{TerraInc} &\textbf{Avg} \\ \midrule
    \multicolumn{6}{c}{train-validation selection criterion} \\ \midrule
    ViT & 89.1$\scalebox{0.8}{$\pm$0.0}$ &  \textbf{80.7$\scalebox{0.8}{$\pm$0.1}$} & 77.2$\scalebox{0.8}{$\pm$0.1}$ & 47.3$\scalebox{0.8}{$\pm$0.8}$ & 73.58 \\
    GMoE &  \textbf{90.0$\scalebox{0.8}{$\pm$0.3}$} & 80.4$\scalebox{0.8}{$\pm$0.6}$ & 77.0$\scalebox{0.8}{$\pm$0.1}$ & \textbf{49.2$\scalebox{0.8}{$\pm$1.1}$} & \textbf{74.15} \\
   $\scalebox{0.9}{EMoE-learn}$ & 89.8$\scalebox{0.8}{$\pm$0.2}$ & 80.6$\scalebox{0.8}{$\pm$0.2}$ & 76.5$\scalebox{0.8}{$\pm$0.1}$ & 48.7$\scalebox{0.8}{$\pm$0.5}$ & 73.90 \\
    EMoE & 89.4$\scalebox{0.8}{$\pm$0.4}$ & 80.7$\scalebox{0.8}{$\pm$0.2}$ & \textbf{77.3$\scalebox{0.8}{$\pm$0.1}$} & 48.5$\scalebox{0.8}{$\pm$0.5}$ &73.98 \\ \midrule
    \multicolumn{6}{c}{leave-one-domain-out selection criterion} \\ \midrule
    ViT & 88.9$\scalebox{0.8}{$\pm$0.4}$ & 80.8$\scalebox{0.8}{$\pm$0.3}$ & \textbf{77.5$\scalebox{0.8}{$\pm$0.1}$} & 46.1$\scalebox{0.8}{$\pm$0.6}$ &73.33 \\
    GMoE & 89.3$\scalebox{0.8}{$\pm$0.6}$ & 81.0$\scalebox{0.8}{$\pm$0.3}$ & 76.7$\scalebox{0.8}{$\pm$0.1}$ & 50.1$\scalebox{0.8}{$\pm$1.1}$  &74.28\\
   $\scalebox{0.9}{EMoE-learn}$ & 89.3$\scalebox{0.8}{$\pm$0.2}$ & 81.2$\scalebox{0.8}{$\pm$0.1}$ & 76.5$\scalebox{0.8}{$\pm$0.1}$ & \textbf{50.5$\scalebox{0.8}{$\pm$0.2}$} &74.38 \\
    EMoE & \textbf{89.6$\scalebox{0.8}{$\pm$0.2}$} & \textbf{81.6$\scalebox{0.8}{$\pm$0.2}$} & 77.4$\scalebox{0.8}{$\pm$0.1}$ & 50.0$\scalebox{0.8}{$\pm$1.1}$  &\textbf{74.65}\\ \midrule
    \multicolumn{6}{c}{test-domain selection criterion} \\ \midrule
    ViT & 88.8$\scalebox{0.8}{$\pm$0.7}$ & 79.0$\scalebox{0.8}{$\pm$0.5}$ & 77.2$\scalebox{0.8}{$\pm$0.0}$ & 46.7$\scalebox{0.8}{$\pm$0.4}$ &72.93\\
    GMoE & 89.7$\scalebox{0.8}{$\pm$0.5}$ & 79.0$\scalebox{0.8}{$\pm$0.3}$ & 77.0$\scalebox{0.8}{$\pm$0.1}$ & \textbf{48.8$\scalebox{0.8}{$\pm$0.4}$} &73.63\\
   $\scalebox{0.9}{EMoE-learn}$ & \textbf{89.7$\scalebox{0.8}{$\pm$0.4}$} & \textbf{79.7$\scalebox{0.8}{$\pm$0.2}$} & 76.6$\scalebox{0.8}{$\pm$0.1}$ & 48.7$\scalebox{0.8}{$\pm$0.3}$ &73.68\\
    EMoE & \textbf{89.7$\scalebox{0.8}{$\pm$0.4}$} & \textbf{79.7$\scalebox{0.8}{$\pm$0.2}$} & \textbf{77.5$\scalebox{0.8}{$\pm$0.1}$} & 48.8$\scalebox{0.8}{$\pm$0.6}$ &\textbf{73.93}\\
    \bottomrule
    \end{tabular}%
    }
  \end{minipage}
\end{table*}

\begin{table*}[h]
  \setlength{\tabcolsep}{4pt}
  \renewcommand{\arraystretch}{1.0}
  \centering
  \caption{Results on GLUE tasks with different backbones and algorithms. All the reported results are obtained from 3 independent experiments. The average accuracy (avg) is reported along with the relative improvement compared to the baseline. The best result is highlighted in \textbf{bold}.}
  \label{tab:results_full_tuning_language_with_std}
  \resizebox{0.8\textwidth}{!}{
  \begin{tabular}{c|l|cccccc}
  \toprule
  \textbf{$\scalebox{0.8}{Backbone}$} & \textbf{$\scalebox{0.8}{Algorithm}$} & \textbf{MRPC} & \textbf{CoLA} & \textbf{RTE} & \textbf{STSB} & \textbf{SST2} & \textbf{Avg} \\
  \midrule
  \multirow{5}{*}{$\scalebox{0.9}{BERT-B}$} & baseline & 88.45$\scalebox{0.8}{$\pm$0.40}$ & 60.67$\scalebox{0.8}{$\pm$0.54}$ & 68.95$\scalebox{0.8}{$\pm$0.69}$ & 87.87$\scalebox{0.8}{$\pm$0.12}$ & 91.97$\scalebox{0.8}{$\pm$0.19}$ & 79.582 \\
  &$\scalebox{0.9}{noisy tuning}$ & 88.43$\scalebox{0.8}{$\pm$0.12}$ & 61.79$\scalebox{0.8}{$\pm$0.16}$ & 71.36$\scalebox{0.8}{$\pm$0.17}$ & 88.27$\scalebox{0.8}{$\pm$0.94}$ & 92.32$\scalebox{0.8}{$\pm$0.25}$ & 80.43(+0.85)\\
  & GMoE & 88.63$\scalebox{0.8}{$\pm$0.53}$ & 61.25$\scalebox{0.8}{$\pm$2.36}$ & 70.28$\scalebox{0.8}{$\pm$0.68}$ & 88.63$\scalebox{0.8}{$\pm$0.65}$ & 92.28$\scalebox{0.8}{$\pm$0.24}$ & 80.21(+0.63)\\
  &$\scalebox{0.9}{EMoE-learn}$ & 89.05$\scalebox{0.8}{$\pm$0.23}$ & \textbf{62.46$\scalebox{0.8}{$\pm$1.01}$} & \textbf{70.40$\scalebox{0.8}{$\pm$1.28}$} & 88.47$\scalebox{0.8}{$\pm$0.74}$ & 92.58$\scalebox{0.8}{$\pm$0.14}$ & \textbf{80.59(+1.01)} \\
  & EMoE & \textbf{89.45$\scalebox{0.8}{$\pm$0.36}$} & 61.55$\scalebox{0.8}{$\pm$0.67}$ & 69.68$\scalebox{0.8}{$\pm$1.02}$ & \textbf{88.71$\scalebox{0.8}{$\pm$0.50}$} & \textbf{92.89$\scalebox{0.8}{$\pm$0.19}$} & 80.46(+0.87)\\
  \midrule
  \multirow{5}{*}{$\scalebox{0.9}{BERT-L}$} & baseline & 89.82$\scalebox{0.8}{$\pm$1.30}$ & 65.41$\scalebox{0.8}{$\pm$0.47}$ & 74.89$\scalebox{0.8}{$\pm$1.39}$ & 89.87$\scalebox{0.8}{$\pm$0.28}$ & 93.50$\scalebox{0.8}{$\pm$0.24}$ & 82.70 \\
  &$\scalebox{0.9}{noisy tuning}$ & 90.42$\scalebox{0.8}{$\pm$0.35}$ & 64.75$\scalebox{0.8}{$\pm$1.31}$ & 73.41$\scalebox{0.8}{$\pm$1.62}$ & 90.05$\scalebox{0.8}{$\pm$0.46}$ & 93.65$\scalebox{0.8}{$\pm$0.11}$ & 82.46(-0.24)\\
  & GMoE & \textbf{91.24$\scalebox{0.8}{$\pm$0.25}$} & 64.90$\scalebox{0.8}{$\pm$1.26}$ & 74.24$\scalebox{0.8}{$\pm$1.04}$ & 90.00$\scalebox{0.8}{$\pm$0.64}$ & 93.58$\scalebox{0.8}{$\pm$0.25}$ & 82.79(+0.09)\\
  &$\scalebox{0.9}{EMoE-learn}$ & 90.57$\scalebox{0.8}{$\pm$0.43}$ & 65.51$\scalebox{0.8}{$\pm$0.32}$ & 74.72$\scalebox{0.8}{$\pm$2.13}$ & 90.22$\scalebox{0.8}{$\pm$0.49}$ & \textbf{93.73$\scalebox{0.8}{$\pm$0.35}$} & 82.95(+0.25)\\
  & EMoE & 90.74$\scalebox{0.8}{$\pm$0.65}$ & \textbf{65.79$\scalebox{0.8}{$\pm$1.16}$} & \textbf{76.17$\scalebox{0.8}{$\pm$0.00}$} & \textbf{90.31$\scalebox{0.8}{$\pm$0.43}$} & 93.58$\scalebox{0.8}{$\pm$0.32}$ & \textbf{83.32(+0.62)}\\
  \midrule
  \multirow{5}{*}{$\scalebox{0.9}{GPT2}$} & baseline & 84.46$\scalebox{0.8}{$\pm$0.51}$ & 47.07$\scalebox{0.8}{$\pm$1.60}$ & 67.15$\scalebox{0.9}{$\pm$0.51}$ & 86.29$\scalebox{0.8}{$\pm$0.29}$ & 92.13$\scalebox{0.8}{$\pm$0.30}$ & 75.42 \\
  &$\scalebox{0.9}{noisy tuning}$ & 84.15$\scalebox{0.8}{$\pm$0.92}$ & 46.16$\scalebox{0.8}{$\pm$2.79}$ & 67.51$\scalebox{0.8}{$\pm$0.78}$ & 86.09$\scalebox{0.8}{$\pm$0.38}$ & 92.13$\scalebox{0.8}{$\pm$0.27}$ & 75.21(-0.21)\\
  & GMoE & 85.07$\scalebox{0.8}{$\pm$0.45}$ & 47.77$\scalebox{0.8}{$\pm$3.20}$ & 67.51$\scalebox{0.8}{$\pm$0.51}$ & 86.57$\scalebox{0.8}{$\pm$0.29}$ & 92.35$\scalebox{0.8}{$\pm$0.35}$ & 75.85(+0.43)\\
  &$\scalebox{0.8}{EMoE-learn}$ & \textbf{85.73$\scalebox{0.8}{$\pm$0.09}$} & 47.24$\scalebox{0.8}{$\pm$1.48}$ & 67.99$\scalebox{0.8}{$\pm$0.17}$ & \textbf{86.66$\scalebox{0.8}{$\pm$0.32}$} & 92.35$\scalebox{0.8}{$\pm$0.38}$ & 75.99(+0.57)\\
  & EMoE & 85.40$\scalebox{0.8}{$\pm$0.77}$ & \textbf{48.00$\scalebox{0.8}{$\pm$1.50}$} & \textbf{68.95$\scalebox{0.8}{$\pm$0.29}$} & 86.64$\scalebox{0.8}{$\pm$0.16}$ & \textbf{92.70$\scalebox{0.8}{$\pm$0.22}$} & \textbf{76.34(+0.92)}\\
  \bottomrule
  \end{tabular}
  }
\end{table*}

\begin{table*}[h]
  \setlength{\tabcolsep}{4pt}
  \renewcommand{\arraystretch}{1.0}
  \centering
  \caption{Results on various algorithms with different models and tasks. All the reported results are obtained from 3 independent experiments. OOD Metrics (averaged over 14 OOD tasks, lower is better) provide additional information for out-of-distribution generalization. The best result is highlighted in \textbf{bold}, and the second is marked with \underline{underline}.}
  \label{tab:full_results_lora_ID_OOD_std}
  \resizebox{\textwidth}{!}{
  \begin{tabular}{l|cccccccccc}
  \toprule
  \textbf{Algorithm} & \textbf{MRPC} & \textbf{CoLA} & \textbf{RTE} & \textbf{STSB} & \textbf{SST2} & \textbf{QNLI} & \textbf{QQP} & \textbf{MNLI} & \textbf{ID-Avg} & \textbf{$\scalebox{0.8}{OOD}$} \\
  \midrule
  \multicolumn{11}{c}{BERT-Large (340 Million Parameters) Results} \\
  \midrule
  LoRA & 89.97$\scalebox{0.8}{$\pm$0.40}$ & 63.40$\scalebox{0.8}{$\pm$0.62}$ & 72.92$\scalebox{0.8}{$\pm$1.64}$ & \underline{90.51$\scalebox{0.8}{$\pm$0.18}$} & 93.16$\scalebox{0.8}{$\pm$0.19}$ & 92.20 $\scalebox{0.8}{$\pm$0.13}$ & 87.21$\scalebox{0.8}{$\pm$0.60}$ & 85.40$\scalebox{0.8}{$\pm$0.07}$ & 84.35 & 4.86 \\
  Block & 89.34$\scalebox{0.8}{$\pm$0.84}$ & 62.10$\scalebox{0.8}{$\pm$0.91}$ & 71.96$\scalebox{0.8}{$\pm$1.68}$ & 90.39$\scalebox{0.8}{$\pm$0.14}$ & 93.35$\scalebox{0.8}{$\pm$0.43}$ & 92.04$\scalebox{0.8}{$\pm$0.16}$ & \underline{88.45$\scalebox{0.8}{$\pm$0.07}$} & \underline{86.20$\scalebox{0.8}{$\pm$0.10}$} & 84.23(-0.12)& 4.95 \\
  Block+GMoE & 89.45$\scalebox{0.8}{$\pm$0.72}$ & 63.80$\scalebox{0.8}{$\pm$0.71}$ & 72.56$\scalebox{0.8}{$\pm$0.29}$ & 90.29$\scalebox{0.8}{$\pm$0.07}$ & \textbf{93.85$\scalebox{0.8}{$\pm$0.11}$} & 92.32$\scalebox{0.8}{$\pm$0.14}$ & 87.99$\scalebox{0.8}{$\pm$0.06}$ & 85.92$\scalebox{0.8}{$\pm$0.13}$ & 84.52(+0.18)& \underline{4.04} \\
 $\scalebox{0.8}{Block+EMoE-learn}$ & 89.79$\scalebox{0.8}{$\pm$0.23}$ & 64.16$\scalebox{0.8}{$\pm$0.87}$ & 73.16$\scalebox{0.8}{$\pm$1.04}$ & 90.27$\scalebox{0.8}{$\pm$0.03}$ & \textbf{93.85$\scalebox{0.8}{$\pm$0.11}$} & \underline{92.40$\scalebox{0.8}{$\pm$0.06}$} & 88.01$\scalebox{0.8}{$\pm$0.12}$ & 85.76$\scalebox{0.8}{$\pm$0.19}$ & 84.68(+0.33) & \textbf{3.94} \\
  Block+EMoE & 89.77$\scalebox{0.8}{$\pm$0.46}$ & 63.25$\scalebox{0.8}{$\pm$0.50}$ & 71.60$\scalebox{0.8}{$\pm$0.68}$ & 90.31$\scalebox{0.8}{$\pm$0.09}$ & 93.69$\scalebox{0.8}{$\pm$0.32}$ & 92.09$\scalebox{0.8}{$\pm$0.13}$ & 88.08$\scalebox{0.8}{$\pm$0.19}$ & \textbf{86.21$\scalebox{0.8}{$\pm$0.16}$} & 84.38(+0.03)& 5.89 \\
  EMoE & \textbf{90.85$\scalebox{0.8}{$\pm$0.61}$} & \textbf{65.33$\scalebox{0.8}{$\pm$0.40}$} & \textbf{75.21$\scalebox{0.8}{$\pm$1.62}$} & 90.43$\scalebox{0.8}{$\pm$0.06}$ & 93.50$\scalebox{0.8}{$\pm$0.33}$ & 92.23$\scalebox{0.8}{$\pm$0.10}$ & 87.74$\scalebox{0.8}{$\pm$0.10}$ & 85.43$\scalebox{0.8}{$\pm$0.10}$ & \textbf{85.09(+0.74)}& 4.37 \\
  EMoE+LN & \underline{90.47$\scalebox{0.8}{$\pm$0.33}$} & \underline{64.39$\scalebox{0.8}{$\pm$0.31}$} & \underline{73.41$\scalebox{0.8}{$\pm$1.04}$} & \textbf{90.54$\scalebox{0.8}{$\pm$0.03}$} & 93.00$\scalebox{0.8}{$\pm$0.16}$ & 92.31$\scalebox{0.8}{$\pm$0.05}$ & \textbf{88.79$\scalebox{0.8}{$\pm$0.17}$} & 85.50$\scalebox{0.8}{$\pm$0.10}$ & \underline{84.80(+0.46)}& 4.53 \\
  EMoE-learn & 89.87$\scalebox{0.8}{$\pm$0.50}$ & 64.00$\scalebox{0.8}{$\pm$0.57}$ & 71.36$\scalebox{0.8}{$\pm$1.39}$ & 90.48$\scalebox{0.8}{$\pm$0.10}$ & 93.65$\scalebox{0.8}{$\pm$0.33}$ & \textbf{92.40$\scalebox{0.8}{$\pm$0.11}$} & 87.55$\scalebox{0.8}{$\pm$0.14}$ & 85.62$\scalebox{0.8}{$\pm$0.23}$ & 84.37(+0.02)& 4.66 \\
 $\scalebox{0.8}{EMoE-learn+LN}$ & 89.9$\scalebox{0.8}{$\pm$0.25}$ & 64.16$\scalebox{0.8}{$\pm$1.16}$ & 72.44$\scalebox{0.8}{$\pm$0.45}$ & 90.45$\scalebox{0.8}{$\pm$0.10}$ & 93.42$\scalebox{0.8}{$\pm$0.38}$ & 92.15$\scalebox{0.8}{$\pm$0.10}$ & 87.70$\scalebox{0.8}{$\pm$0.04}$ & 85.52$\scalebox{0.8}{$\pm$0.24}$ & 84.47(0.12)& 4.28 \\
  \midrule
  \multicolumn{11}{c}{GPT2-XL (1.5 Billion Parameters) Results} \\
  \midrule
  LoRA & 86.83$\scalebox{0.8}{$\pm$0.87}$ & 60.88$\scalebox{0.8}{$\pm$2.54}$ & 78.70$\scalebox{0.8}{$\pm$0.59}$ & 89.07$\scalebox{0.8}{$\pm$0.11}$ & 95.18$\scalebox{0.8}{$\pm$0.28}$ & 91.84$\scalebox{0.8}{$\pm$0.09}$ & 87.41$\scalebox{0.8}{$\pm$1.74}$ & 86.93$\scalebox{0.8}{$\pm$0.15}$ & 84.61 & 5.61 \\
  Block & 86.59$\scalebox{0.8}{$\pm$1.45}$ & 61.18$\scalebox{0.8}{$\pm$1.74}$ & 79.78$\scalebox{0.8}{$\pm$2.22}$ & 89.08$\scalebox{0.8}{$\pm$0.15}$ & \underline{95.45$\scalebox{0.8}{$\pm$0.19}$} & 91.88$\scalebox{0.8}{$\pm$0.05}$ & 87.71$\scalebox{0.8}{$\pm$2.95}$ & 86.95$\scalebox{0.8}{$\pm$0.08}$ & 84.83(+0.22)& 5.13 \\
  Block+GMoE & 87.02$\scalebox{0.8}{$\pm$0.76}$ & 62.81$\scalebox{0.8}{$\pm$1.51}$ & 79.78$\scalebox{0.8}{$\pm$1.35}$ & 89.21$\scalebox{0.8}{$\pm$0.20}$ & 95.41$\scalebox{0.8}{$\pm$0.28}$ & 92.18$\scalebox{0.8}{$\pm$0.11}$ & 89.10$\scalebox{0.8}{$\pm$0.78}$ & \textbf{87.17$\scalebox{0.8}{$\pm$0.20}$} & 85.34(+0.73)& 4.33 \\
 $\scalebox{0.8}{Block+EMoE-learn}$ & 87.31$\scalebox{0.8}{$\pm$1.23}$ & 62.24$\scalebox{0.8}{$\pm$1.51}$ & 79.54$\scalebox{0.8}{$\pm$0.17}$ & 89.33$\scalebox{0.8}{$\pm$0.11}$ & 95.30$\scalebox{0.8}{$\pm$0.09}$ & 92.20$\scalebox{0.8}{$\pm$0.09}$ & 88.59$\scalebox{0.8}{$\pm$1.68}$ & \underline{87.06$\scalebox{0.8}{$\pm$0.18}$} & 85.20(+0.59)& 4.05 \\
  Block+EMoE & 87.86$\scalebox{0.8}{$\pm$0.98}$ & \underline{62.88$\scalebox{0.8}{$\pm$0.54}$} & \textbf{80.05$\scalebox{0.8}{$\pm$0.29}$} & 89.18$\scalebox{0.8}{$\pm$0.25}$ & \textbf{95.49$\scalebox{0.8}{$\pm$0.39}$} & 92.10$\scalebox{0.8}{$\pm$0.15}$ & 89.69$\scalebox{0.8}{$\pm$0.15}$ & 86.87$\scalebox{0.8}{$\pm$0.11}$ & \underline{85.52(+0.91)}& 5.71 \\
  EMoE & 87.75$\scalebox{0.8}{$\pm$0.14}$ & 62.27$\scalebox{0.8}{$\pm$0.93}$ & \underline{80.02$\scalebox{0.8}{$\pm$0.34}$} & 89.37$\scalebox{0.8}{$\pm$0.30}$ & 95.41$\scalebox{0.8}{$\pm$0.32}$ & 92.10$\scalebox{0.8}{$\pm$0.15}$ & 89.58$\scalebox{0.8}{$\pm$0.10}$ & 87.06$\scalebox{0.8}{$\pm$0.25}$ & 85.45(+0.84)& \underline{3.88} \\
  EMoE+LN & \textbf{88.05$\scalebox{0.8}{$\pm$0.35}$} & \textbf{63.11$\scalebox{0.8}{$\pm$0.51}$} & 79.90$\scalebox{0.8}{$\pm$1.51}$ & 89.40$\scalebox{0.8}{$\pm$0.22}$ & 95.18$\scalebox{0.8}{$\pm$0.28}$ & 92.23$\scalebox{0.8}{$\pm$0.11}$ & \underline{89.70}$\scalebox{0.8}{$\pm$0.09}$ & 87.03$\scalebox{0.8}{$\pm$0.14}$ & \textbf{85.58(+0.97)}& 4.39 \\
  EMoE-learn & \underline{87.93$\scalebox{0.8}{$\pm$0.61}$} & 61.50$\scalebox{0.8}{$\pm$1.09}$ & 79.90$\scalebox{0.8}{$\pm$0.61}$ & \underline{89.48$\scalebox{0.8}{$\pm$0.24}$} & 95.18$\scalebox{0.8}{$\pm$0.11}$ & \textbf{92.33$\scalebox{0.8}{$\pm$0.093}$} & \textbf{89.71$\scalebox{0.8}{$\pm$0.06}$} & 87.00$\scalebox{0.8}{$\pm$0.19}$ & 85.38(+0.77)& 4.40 \\
 $\scalebox{0.8}{EMoE-learn+LN}$ & 87.04$\scalebox{0.8}{$\pm$1.11}$ & 62.64$\scalebox{0.8}{$\pm$0.84}$ & 79.78$\scalebox{0.8}{$\pm$0.59}$ & \textbf{89.50$\scalebox{0.8}{$\pm$0.22}$} & 95.30$\scalebox{0.8}{$\pm$0.50}$ & \underline{92.31$\scalebox{0.8}{$\pm$0.19}$} & 89.43$\scalebox{0.8}{$\pm$0.35}$ & 87.00$\scalebox{0.8}{$\pm$0.12}$ & 85.38(+0.77)& \textbf{3.67} \\
  \bottomrule
  \end{tabular}
  }
\end{table*}

\subsection{Full Tables and Figure Data Sources in Analysis}
\label{appdendix:full_analysis_results}
In the Analysis section~\ref{sec:analysis}, we have transformed tabular data into graphs or retained only a subset of the results for clarity. 
The original and complete results corresponding to them are presented in this section.
Fig.~\ref{fig:results-analysis} is summarized from Tab.~\ref{tab:results_compare_train_test_std} and Tab.~\ref{tab:full_results_selections}. 
The OOD results in Table~\ref{tab:results_lora_ID_OOD} are from \ref{tab:full_results_lora_ID_OOD_std}. 
The ablation studies are from Tab.~\ref{tab:results_appendix_more_MoEs_config},  and Tab.\ref{tab:results_comparison_datafraction}.

\begin{table*}[h]
  \setlength{\tabcolsep}{4pt}
  \renewcommand{\arraystretch}{1.0}
  \centering
  \caption{ID and OOD results of BERT-L for different settings. "LoRA-to-EMoE" refers to converting a model tuned using standard LoRA into EMoE for testing. On the other hand, "EMoE-to-LoRA" involves merging a tuned EMoE model back into a standard standard model during testing.}
  \label{tab:results_compare_train_test_std}
  \resizebox{\textwidth}{!}{
    \begin{tabular}{l|ccccccccccc}
      \toprule
      \textbf{Algorithm} & \textbf{CoLA} & \textbf{Gram} & \textbf{MRPC} & \textbf{Twitter} & \textbf{RTE} & \textbf{Hans} & \textbf{SciTail} & \textbf{STSB} & \textbf{Sick} & \textbf{Avg} \\
      \midrule
      LoRA & 60.89$\scalebox{0.8}{$\pm$2.55}$ & 41.77$\scalebox{0.8}{$\pm$1.62}$ & 86.83$\scalebox{0.8}{$\pm$0.87}$ & 75.42$\scalebox{0.8}{$\pm$2.71}$ & 78.70$\scalebox{0.8}{$\pm$1.02}$ & 60.37$\scalebox{0.8}{$\pm$1.31}$ & 77.36$\scalebox{0.8}{$\pm$0.73}$ & 89.22$\scalebox{0.8}{$\pm$0.13}$ & 78.48$\scalebox{0.8}{$\pm$0.33}$ & 72.12\\
       $\scalebox{0.8}{LoRA-to-EMoE}$ & 61.31$\scalebox{0.8}{$\pm$2.14}$ & 41.99$\scalebox{0.8}{$\pm$1.56}$ & 86.83$\scalebox{0.8}{$\pm$0.91}$ & 75.15$\scalebox{0.8}{$\pm$3.02}$ & 78.58$\scalebox{0.8}{$\pm$0.95}$ & 60.39$\scalebox{0.8}{$\pm$1.32}$ & 77.24$\scalebox{0.8}{$\pm$0.68}$ & 89.23$\scalebox{0.8}{$\pm$0.13}$ & 78.53$\scalebox{0.8}{$\pm$0.35}$ & 72.14\\
      EMoE & 62.69$\scalebox{0.8}{$\pm$0.91}$ & 42.95$\scalebox{0.8}{$\pm$0.95}$ & 87.82$\scalebox{0.8}{$\pm$0.17}$ & 76.07$\scalebox{0.8}{$\pm$2.12}$ & 79.54$\scalebox{0.8}{$\pm$0.45}$ & 61.56$\scalebox{0.8}{$\pm$1.65}$ & 78.09$\scalebox{0.8}{$\pm$0.56}$ & 89.39$\scalebox{0.8}{$\pm$0.31}$ & 78.57$\scalebox{0.8}{$\pm$0.67}$ & 72.96\\
      $\scalebox{0.8}{EMoE-to-LoRA}$  & 62.69$\scalebox{0.8}{$\pm$0.91}$ & 42.94$\scalebox{0.8}{$\pm$0.95}$ & 87.82$\scalebox{0.8}{$\pm$0.17}$ & 76.06$\scalebox{0.8}{$\pm$2.12}$ & 79.54$\scalebox{0.8}{$\pm$0.45}$ & 61.56$\scalebox{0.8}{$\pm$1.65}$ & 78.07$\scalebox{0.8}{$\pm$0.58}$ & 89.39$\scalebox{0.8}{$\pm$0.3}$ & 78.58$\scalebox{0.8}{$\pm$0.67}$ & 72.96\\
      \bottomrule
    \end{tabular}
  }
\end{table*}

\begin{table*}[h]
  \setlength{\tabcolsep}{4pt}
  \renewcommand{\arraystretch}{1.0}
  \centering
  \caption{ID results of BERT-L for different settings. "Cluster-top" refers to EMoE utilizing $\operatorname{avg-k}$ gating. "Cluster-not-top" represents a scenario where, during gating, the top-k experts are removed. Similarly, "Cluster-bottom" involves selecting the bottom-k experts with the lowest scores during gating. "Random" denotes the approach of randomly selecting key values to construct experts. The terms "top," "not-top," and "bottom" have the same meanings as in the cluster situations.}
  \label{tab:full_results_selections}
  \resizebox{0.9\textwidth}{!}{
    \begin{tabular}{l|cccccccc}
      \toprule
      \textbf{Algorithm} & \textbf{MRPC} & \textbf{CoLA} & \textbf{RTE} & \textbf{STSB} & \textbf{SST2} & \textbf{QNLI} & \textbf{Avg} \\
      \midrule
      LoRA & 89.97$\scalebox{0.8}{$\pm$0.40}$ & 63.40$\scalebox{0.8}{$\pm$0.62}$ & 72.92$\scalebox{0.8}{$\pm$1.64}$ & 90.51$\scalebox{0.8}{$\pm$0.18}$ & 93.16$\scalebox{0.8}{$\pm$0.19}$ & 92.20$\scalebox{0.8}{$\pm$0.13}$ & 83.69 \\
      Cluster-top & 90.85$\scalebox{0.8}{$\pm$0.61}$ & 65.33$\scalebox{0.8}{$\pm$0.40}$ & 75.21$\scalebox{0.8}{$\pm$1.62}$ & 90.54$\scalebox{0.8}{$\pm$0.03}$ & 93.50$\scalebox{0.8}{$\pm$0.33}$ & 92.23$\scalebox{0.8}{$\pm$0.10}$ & 84.61(+0.92) \\
      Cluster-not-top & 89.61$\scalebox{0.8}{$\pm$0.76}$ & 63.21$\scalebox{0.8}{$\pm$0.44}$ & 72.56$\scalebox{0.8}{$\pm$1.28}$ & 90.31$\scalebox{0.8}{$\pm$0.07}$ & 93.12$\scalebox{0.8}{$\pm$0.34}$ & 92.14$\scalebox{0.8}{$\pm$0.18}$ & 83.49(-0.20) \\
      Cluster-bottom & 89.21$\scalebox{0.8}{$\pm$0.69}$ & 63.08$\scalebox{0.8}{$\pm$1.09}$ & 71.72$\scalebox{0.8}{$\pm$0.34}$ & 90.15$\scalebox{0.8}{$\pm$0.18}$ & 92.97$\scalebox{0.8}{$\pm$0.19}$ & 92.13$\scalebox{0.8}{$\pm$0.31}$ & 83.21(-0.48) \\
      Random-top & 89.88$\scalebox{0.8}{$\pm$0.75}$ & 63.26$\scalebox{0.8}{$\pm$0.39}$ & 72.56$\scalebox{0.8}{$\pm$1.06}$ & 90.33$\scalebox{0.8}{$\pm$0.05}$ & 93.35$\scalebox{0.8}{$\pm$0.00}$ & 92.14$\scalebox{0.8}{$\pm$0.20}$ & 83.59(-0.11) \\
      Random-not-top & 90.09$\scalebox{0.8}{$\pm$0.75}$ & 63.35$\scalebox{0.8}{$\pm$0.34}$ & 72.44$\scalebox{0.8}{$\pm$1.19}$ & 90.44$\scalebox{0.8}{$\pm$0.07}$ & 93.31$\scalebox{0.8}{$\pm$0.25}$ & 92.20$\scalebox{0.8}{$\pm$0.16}$ & 83.64(-0.05) \\
      Random-bottom & 89.47$\scalebox{0.8}{$\pm$0.23}$ & 63.17$\scalebox{0.8}{$\pm$0.98}$ & 71.96$\scalebox{0.8}{$\pm$0.74}$ & 90.30$\scalebox{0.8}{$\pm$0.23}$ & 93.11$\scalebox{0.8}{$\pm$0.25}$ & 92.10$\scalebox{0.8}{$\pm$0.11}$ & 83.35(-0.34) \\
      \bottomrule
    \end{tabular}
  }
\end{table*}

\begin{table*}[!ht]
  \vspace{-3em}
    \caption{Raw OOD performances across 13 tasks. Average results with standard deviation and best results are reported separately. Due to the large deviation across seeds overall methods, we use Friedman rank metrics~\citep{friedman1940comparison}.}
      \vspace{-1em}
    \centering
    \rotatebox{-90}{\resizebox{1.4\textwidth}{!}{
  \begin{tabular}{c|ccccccccccccccccccccccccccc}
    \toprule
    \textbf{Algorithm} & \textbf{$\scalebox{0.8}{Twitter-M}$} & \textbf{$\scalebox{0.7}{GrammarTest}$} & \textbf{Hans} & \textbf{SciTail} & \textbf{Sick-S} & $\scalebox{0.8}{\textbf{NewsQA}}$ & \textbf{Amazon} & \textbf{Flipkart} & \textbf{Imdb} & \textbf{Yelp} & \textbf{MRPC} & \textbf{$\scalebox{0.8}{Twitter-Q}$} & \textbf{Sick-M} \\
    \midrule
    \multicolumn{14}{c}{BERT-Large (340 Million Parameters) Results (mean and standard deviation)} \\
    \midrule
  LoRA & 80.09$\scalebox{0.8}{$\pm$0.49}$ & 43.55$\scalebox{0.8}{$\pm$1.34}$ & 55.51$\scalebox{0.8}{$\pm$1.94}$ & 81.13$\scalebox{0.8}{$\pm$0.88}$ & 81.63$\scalebox{0.8}{$\pm$0.16}$ & 78.22$\scalebox{0.8}{$\pm$0.17}$ & 89.06$\scalebox{0.8}{$\pm$1.81}$ & 91.79$\scalebox{0.8}{$\pm$0.27}$ & 84.97$\scalebox{0.8}{$\pm$1.32}$ & 88.40$\scalebox{0.8}{$\pm$2.06}$ & 71.73$\scalebox{0.8}{$\pm$0.46}$ & 78.10$\scalebox{0.8}{$\pm$0.33}$ & 53.17$\scalebox{0.8}{$\pm$0.92}$ \\
  LoRA+block & 80.70$\scalebox{0.8}{$\pm$0.35}$ & 42.94$\scalebox{0.8}{$\pm$0.47}$ & 54.35$\scalebox{0.8}{$\pm$1.08}$ & 80.99$\scalebox{0.8}{$\pm$0.33}$ & 81.58$\scalebox{0.8}{$\pm$0.18}$ & 78.33$\scalebox{0.8}{$\pm$0.24}$ & 88.21$\scalebox{0.8}{$\pm$1.20}$ & 91.26$\scalebox{0.8}{$\pm$0.96}$ & 84.10$\scalebox{0.8}{$\pm$0.85}$ & 89.13$\scalebox{0.8}{$\pm$0.20}$ & 71.57$\scalebox{0.8}{$\pm$0.87}$ & 79.22$\scalebox{0.8}{$\pm$0.59}$ & 53.67$\scalebox{0.8}{$\pm$0.77}$ \\
  GMoE & 81.17$\scalebox{0.8}{$\pm$0.54}$ & 46.76$\scalebox{0.8}{$\pm$1.91}$ & 58.73$\scalebox{0.8}{$\pm$0.74}$ & 79.41$\scalebox{0.8}{$\pm$0.10}$ & 81.35$\scalebox{0.8}{$\pm$0.34}$ & 78.05$\scalebox{0.8}{$\pm$0.28}$ & 89.27$\scalebox{0.8}{$\pm$1.38}$ & 91.77$\scalebox{0.8}{$\pm$0.07}$ & 84.93$\scalebox{0.8}{$\pm$1.09}$ & 88.70$\scalebox{0.8}{$\pm$1.77}$ & 71.08$\scalebox{0.8}{$\pm$0.69}$ & 78.70$\scalebox{0.8}{$\pm$0.37}$ & 53.85$\scalebox{0.8}{$\pm$1.27}$ \\
  $\scalebox{0.8}{Block+EMoE-learn}$ & 81.09$\scalebox{0.8}{$\pm$0.47}$ & 43.29$\scalebox{0.8}{$\pm$1.42}$ & 57.55$\scalebox{0.8}{$\pm$3.48}$ & 77.72$\scalebox{0.8}{$\pm$0.61}$ & 80.78$\scalebox{0.8}{$\pm$0.49}$ & 78.32$\scalebox{0.8}{$\pm$0.13}$ & 90.17$\scalebox{0.8}{$\pm$0.32}$ & 91.82$\scalebox{0.8}{$\pm$0.18}$ & 85.71$\scalebox{0.8}{$\pm$0.19}$ & 89.62$\scalebox{0.8}{$\pm$0.24}$ & 72.71$\scalebox{0.8}{$\pm$0.23}$ & 77.85$\scalebox{0.8}{$\pm$1.03}$ & 54.69$\scalebox{0.8}{$\pm$0.15}$ \\
  Block+EMoE & 81.07$\scalebox{0.8}{$\pm$0.53}$ & 46.33$\scalebox{0.8}{$\pm$1.40}$ & 53.50$\scalebox{0.8}{$\pm$0.97}$ & 79.62$\scalebox{0.8}{$\pm$0.45}$ & 80.15$\scalebox{0.8}{$\pm$0.61}$ & 79.03$\scalebox{0.8}{$\pm$0.25}$ & 88.81$\scalebox{0.8}{$\pm$0.19}$ & 91.33$\scalebox{0.8}{$\pm$0.23}$ & 84.92$\scalebox{0.8}{$\pm$0.07}$ & 88.87$\scalebox{0.8}{$\pm$0.26}$ & 70.59$\scalebox{0.8}{$\pm$1.00}$ & 74.62$\scalebox{0.8}{$\pm$2.70}$ & 53.04$\scalebox{0.8}{$\pm$1.82}$ \\
  EMoE & 80.82$\scalebox{0.8}{$\pm$0.10}$ & 44.18$\scalebox{0.8}{$\pm$0.63}$ & 61.30$\scalebox{0.8}{$\pm$2.24}$ & 76.93$\scalebox{0.8}{$\pm$2.52}$ & 81.19$\scalebox{0.8}{$\pm$0.27}$ & 77.87$\scalebox{0.8}{$\pm$0.19}$ & 90.25$\scalebox{0.8}{$\pm$0.10}$ & 92.14$\scalebox{0.8}{$\pm$0.25}$ & 85.63$\scalebox{0.8}{$\pm$0.30}$ & 90.21$\scalebox{0.8}{$\pm$0.14}$ & 71.24$\scalebox{0.8}{$\pm$1.22}$ & 78.00$\scalebox{0.8}{$\pm$0.87}$ & 51.55$\scalebox{0.8}{$\pm$0.88}$ \\
  EMoE+ln & 81.04$\scalebox{0.8}{$\pm$0.10}$ & 45.05$\scalebox{0.8}{$\pm$0.88}$ & 58.17$\scalebox{0.8}{$\pm$1.81}$ & 77.92$\scalebox{0.8}{$\pm$1.64}$ & 81.27$\scalebox{0.8}{$\pm$0.17}$ & 77.40$\scalebox{0.8}{$\pm$0.25}$ & 90.08$\scalebox{0.8}{$\pm$0.07}$ & 91.64$\scalebox{0.8}{$\pm$0.54}$ & 85.67$\scalebox{0.8}{$\pm$0.29}$ & 89.23$\scalebox{0.8}{$\pm$0.67}$ & 72.14$\scalebox{0.8}{$\pm$0.76}$ & 77.34$\scalebox{0.8}{$\pm$1.33}$ & 52.81$\scalebox{0.8}{$\pm$0.48}$ \\
  EMoE-learn & 81.08$\scalebox{0.8}{$\pm$0.24}$ & 46.09$\scalebox{0.8}{$\pm$2.12}$ & 57.56$\scalebox{0.8}{$\pm$0.43}$ & 77.52$\scalebox{0.8}{$\pm$2.15}$ & 81.25$\scalebox{0.8}{$\pm$0.13}$ & 78.16$\scalebox{0.8}{$\pm$0.30}$ & 89.84$\scalebox{0.8}{$\pm$0.13}$ & 91.87$\scalebox{0.8}{$\pm$0.28}$ & 84.97$\scalebox{0.8}{$\pm$0.33}$ & 89.76$\scalebox{0.8}{$\pm$0.33}$ & 71.32$\scalebox{0.8}{$\pm$2.03}$ & 77.47$\scalebox{0.8}{$\pm$0.76}$ & 53.29$\scalebox{0.8}{$\pm$0.45}$ \\
  $\scalebox{0.8}{EMoE-learn+ln}$ & 81.34$\scalebox{0.8}{$\pm$0.30}$ & 44.46$\scalebox{0.8}{$\pm$1.67}$ & 58.38$\scalebox{0.8}{$\pm$0.59}$ & 77.78$\scalebox{0.8}{$\pm$1.71}$ & 81.13$\scalebox{0.8}{$\pm$0.26}$ & 78.28$\scalebox{0.8}{$\pm$0.13}$ & 90.02$\scalebox{0.8}{$\pm$0.29}$ & 91.48$\scalebox{0.8}{$\pm$0.30}$ & 85.62$\scalebox{0.8}{$\pm$0.46}$ & 89.87$\scalebox{0.8}{$\pm$0.30}$ & 71.90$\scalebox{0.8}{$\pm$0.61}$ & 77.38$\scalebox{0.8}{$\pm$1.77}$ & 53.44$\scalebox{0.8}{$\pm$0.47}$ \\
    \midrule
    \multicolumn{14}{c}{BERT-Large (340 Million Parameters) Results (best result)} \\
    \midrule
    LoRA & 80.75 & 45.42 & 58.24 & 82.03 & 81.79 & 78.46 & 90.35 & 92.05 & 85.95 & 90.37 & 72.06 & 78.41 & 54.44 \\
    LoRA+block & 81.19 & 43.56 & 55.48 & 81.45 & 81.83 & 78.63 & 89.49 & 92.48 & 85.25 & 89.17 & 72.79 & 79.88 & 54.71 \\
    GMoE & 81.78 & 49.14 & 59.72 & 79.55 & 81.66 & 78.41 & 90.31 & 91.86 & 85.93 & 90.42 & 72.06 & 79.02 & 55.16 \\
    $\scalebox{0.8}{Block+EMoE-learn}$ & 81.72 & 44.45 & 61.70 & 78.47 & 81.38 & 78.46 & 90.59 & 92.04 & 85.93 & 89.95 & 73.04 & 78.89 & 54.91 \\
    Block+EMoE & 81.49 & 47.77 & 54.57 & 80.15 & 80.85 & 79.32 & 88.98 & 91.66 & 85.00 & 89.23 & 71.81 & 77.14 & 55.56 \\
    EMoE & 80.95 & 45.05 & 64.31 & 79.64 & 81.55 & 78.10 & 90.37 & 92.42 & 86.03 & 90.34 & 72.55 & 79.23 & 52.24 \\
    EMoE+ln & 81.17 & 45.71 & 60.73 & 79.84 & 81.48 & 77.63 & 90.15 & 92.33 & 86.01 & 90.60 & 72.79 & 79.53 & 53.28 \\
    EMoE-learn & 81.34 & 48.75 & 58.11 & 79.89 & 81.25 & 78.58 & 89.93 & 92.23 & 85.38 & 90.21 & 74.02 & 78.33 & 53.91 \\
    $\scalebox{0.8}{EMoE-learn+ln}$ & 81.72 & 46.82 & 59.09 & 79.60 & 81.41 & 78.47 & 90.27 & 91.72 & 86.11 & 90.17 & 72.55 & 78.67 & 54.10 \\
  
    \midrule
    \multicolumn{14}{c}{GPT2-XL (1.5 Billion Parameters) Results (mean and standard deviation)} \\
    \midrule
    LoRA & 75.42$\scalebox{0.8}{$\pm$2.71}$ & 41.78$\scalebox{0.8}{$\pm$1.62}$ & 60.37$\scalebox{0.8}{$\pm$1.32}$ & 77.36$\scalebox{0.8}{$\pm$0.73}$ & 78.48$\scalebox{0.8}{$\pm$0.33}$ & 78.57$\scalebox{0.8}{$\pm$0.59}$ & 89.91$\scalebox{0.8}{$\pm$0.73}$ & 90.05$\scalebox{0.8}{$\pm$0.44}$ & 85.57$\scalebox{0.8}{$\pm$1.34}$ & 88.85$\scalebox{0.8}{$\pm$0.38}$ & 68.22$\scalebox{0.8}{$\pm$1.27}$ & 73.60$\scalebox{0.8}{$\pm$2.87}$ & 57.39$\scalebox{0.8}{$\pm$0.34}$ \\
    LoRA+block & 76.87$\scalebox{0.8}{$\pm$2.47}$ & 41.09$\scalebox{0.8}{$\pm$1.80}$ & 58.67$\scalebox{0.8}{$\pm$0.97}$ & 78.57$\scalebox{0.8}{$\pm$1.23}$ & 77.66$\scalebox{0.8}{$\pm$0.25}$ & 78.78$\scalebox{0.8}{$\pm$0.55}$ & 90.11$\scalebox{0.8}{$\pm$0.71}$ & 91.54$\scalebox{0.8}{$\pm$0.94}$ & 83.54$\scalebox{0.8}{$\pm$0.57}$ & 88.85$\scalebox{0.8}{$\pm$1.01}$ & 69.53$\scalebox{0.8}{$\pm$0.64}$ & 68.22$\scalebox{0.8}{$\pm$7.84}$ & 58.53$\scalebox{0.8}{$\pm$0.46}$ \\
    GMoE & 75.28$\scalebox{0.8}{$\pm$3.67}$ & 42.50$\scalebox{0.8}{$\pm$1.37}$ & 62.59$\scalebox{0.8}{$\pm$0.37}$ & 77.89$\scalebox{0.8}{$\pm$0.56}$ & 78.33$\scalebox{0.8}{$\pm$0.58}$ & 79.18$\scalebox{0.8}{$\pm$0.04}$ & 90.11$\scalebox{0.8}{$\pm$0.38}$ & 88.64$\scalebox{0.8}{$\pm$4.52}$ & 83.23$\scalebox{0.8}{$\pm$0.40}$ & 89.47$\scalebox{0.8}{$\pm$0.26}$ & 70.10$\scalebox{0.8}{$\pm$1.39}$ & 74.61$\scalebox{0.8}{$\pm$3.71}$ & 58.29$\scalebox{0.8}{$\pm$0.78}$ \\
    $\scalebox{0.8}{Block+EMoE-learn}$ & 74.80$\scalebox{0.8}{$\pm$5.15}$ & 44.15$\scalebox{0.8}{$\pm$1.95}$ & 61.67$\scalebox{0.8}{$\pm$1.18}$ & 78.18$\scalebox{0.8}{$pm$1.04}$ & 78.27$\scalebox{0.8}{$\pm$0.21}$ & 78.97$\scalebox{0.8}{$\pm$0.20}$ & 90.22$\scalebox{0.8}{$\pm$0.59}$ & 91.04$\scalebox{0.8}{$\pm$0.65}$ & 83.82$\scalebox{0.8}{$\pm$0.54}$ & 89.49$\scalebox{0.8}{$\pm$0.67}$ & 69.93$\scalebox{0.8}{$\pm$0.50}$ & 73.41$\scalebox{0.8}{$\pm$3.94}$ & 57.58$\scalebox{0.8}{$\pm$0.57}$ \\
    Block+EMoE & 75.37$\scalebox{0.8}{$\pm$3.04}$ & 39.33$\scalebox{0.8}{$\pm$2.21}$ & 58.32$\scalebox{0.8}{$\pm$2.40}$ & 72.05$\scalebox{0.8}{$\pm$3.04}$ & 77.70$\scalebox{0.8}{$\pm$0.59}$ & 79.32$\scalebox{0.8}{$\pm$0.06}$ & 90.24$\scalebox{0.8}{$\pm$0.36}$ & 91.79$\scalebox{0.8}{$\pm$0.27}$ & 83.79$\scalebox{0.8}{$\pm$0.27}$ & 89.23$\scalebox{0.8}{$\pm$0.38}$ & 63.24$\scalebox{0.8}{$\pm$8.56}$ & 70.60$\scalebox{0.8}{$\pm$2.89}$ & 56.85$\scalebox{0.8}{$\pm$0.57}$ \\
    EMoE & 76.07$\scalebox{0.8}{$\pm$2.12}$ & 42.95$\scalebox{0.8}{$\pm$0.95}$ & 61.56$\scalebox{0.8}{$\pm$1.65}$ & 78.09$\scalebox{0.8}{$\pm$0.56}$ & 78.57$\scalebox{0.8}{$\pm$0.67}$ & 78.87$\scalebox{0.8}{$\pm$0.38}$ & 90.39$\scalebox{0.8}{$\pm$0.55}$ & 91.87$\scalebox{0.8}{$\pm$0.22}$ & 83.55$\scalebox{0.8}{$\pm$1.04}$ & 89.14$\scalebox{0.8}{$\pm$1.39}$ & 67.73$\scalebox{0.8}{$\pm$1.55}$ & 74.06$\scalebox{0.8}{$\pm$4.79}$ & 57.33$\scalebox{0.8}{$\pm$0.59}$ \\
    EMoE+ln & 74.35$\scalebox{0.8}{$\pm$3.50}$ & 42.24$\scalebox{0.8}{$\pm$0.89}$ & 61.88$\scalebox{0.8}{$\pm$2.45}$ & 78.32$\scalebox{0.8}{$\pm$0.95}$ & 78.67$\scalebox{0.8}{$\pm$0.62}$ & 79.02$\scalebox{0.8}{$\pm$0.30}$ & 89.51$\scalebox{0.8}{$\pm$0.90}$ & 91.19$\scalebox{0.8}{$\pm$0.56}$ & 83.12$\scalebox{0.8}{$\pm$1.90}$ & 89.29$\scalebox{0.8}{$\pm$1.09}$ & 69.20$\scalebox{0.8}{$\pm$1.03}$ & 72.74$\scalebox{0.8}{$\pm$5.10}$ & 57.62$\scalebox{0.8}{$\pm$0.02}$ \\
    EMoE-learn & 74.61$\scalebox{0.8}{$\pm$4.75}$ & 40.50$\scalebox{0.8}{$\pm$0.46}$ & 59.50$\scalebox{0.8}{$\pm$3.74}$ & 80.04$\scalebox{0.8}{$\pm$0.46}$ & 78.27$\scalebox{0.8}{$\pm$0.59}$ & 79.06$\scalebox{0.8}{$\pm$0.50}$ & 90.59$\scalebox{0.8}{$\pm$0.66}$ & 91.69$\scalebox{0.8}{$\pm$0.59}$ & 83.23$\scalebox{0.8}{$\pm$1.54}$ & 89.79$\scalebox{0.8}{$\pm$1.35}$ & 67.48$\scalebox{0.8}{$\pm$1.30}$ & 73.62$\scalebox{0.8}{$\pm$1.12}$ & 56.22$\scalebox{0.8}{$\pm$0.15}$ \\
    $\scalebox{0.8}{EMoE-learn+ln}$ & 77.17$\scalebox{0.8}{$\pm$1.58}$ & 42.52$\scalebox{0.8}{$\pm$0.41}$ & 58.90$\scalebox{0.8}{$\pm$4.16}$ & 79.56$\scalebox{0.8}{$\pm$0.81}$ & 78.27$\scalebox{0.8}{$\pm$0.54}$ & 79.19$\scalebox{0.8}{$\pm$0.32}$ & 90.23$\scalebox{0.8}{$\pm$0.86}$ & 91.98$\scalebox{0.8}{$\pm$0.35}$ & 83.31$\scalebox{0.8}{$\pm$1.08}$ & 88.50$\scalebox{0.8}{$\pm$1.92}$ & 68.63$\scalebox{0.8}{$\pm$1.51}$ & 71.26$\scalebox{0.8}{$\pm$3.49}$ & 56.69$\scalebox{0.8}{$\pm$0.37}$ \\
    \midrule
    \multicolumn{14}{c}{GPT2-XL (1.5 Billion Parameters) Results (best result)} \\
    \midrule
    LoRA & 78.17 & 43.93 & 61.77 & 78.39 & 78.87 & 79.12 & 90.52 & 91.03 & 84.81 & 89.39 & 69.12 & 77.48 & 57.79 \\
    LoRA+block & 80.17 & 43.09 & 59.52 & 79.45 & 78.33 & 79.42 & 90.94 & 92.87 & 84.12 & 90.28 & 70.10 & 74.55 & 57.64 \\
    GMoE & 80.23 & 43.96 & 63.05 & 78.68 & 78.00 & 79.33 & 90.57 & 92.27 & 83.69 & 89.81 & 71.08 & 79.79 & 59.36 \\
    $\scalebox{0.8}{Block+EMoE-learn}$& 80.04 & 46.01 & 62.74 & 79.65 & 79.06 & 79.24 & 90.67 & 91.94 & 84.45 & 90.42 & 70.59 & 78.78 & 58.28 \\
    Block+EMoE & 79.02 & 41.55 & 61.64 & 76.30 & 78.33 & 79.40 & 90.68 & 92.00 & 84.12 & 89.60 & 73.31 & 73.31 & 57.64 \\
    EMoE & 77.58 & 43.84 & 63.78 & 78.75 & 79.29 & 79.26 & 91.17 & 92.19 & 84.42 & 90.22 & 69.85 & 80.81 & 58.01 \\
    EMoE+ln & 77.92 & 43.39 & 65.32 & 79.16 & 79.25 & 79.35 & 90.48 & 91.74 & 85.73 & 90.79 & 70.59 & 79.81 & 57.87 \\
    EMoE-learn & 79.35 & 41.15 & 64.56 & 80.40 & 78.76 & 79.76 & 91.32 & 92.47 & 84.49 & 90.77 & 69.12 & 74.77 & 56.42 \\
    $\scalebox{0.8}{EMoE-learn+ln}$ & 79.35 & 43.01 & 64.32 & 80.57 & 79.02 & 79.63 & 91.40 & 92.41 & 84.84 & 90.57 & 70.34 & 74.77 & 57.13 \\
    \bottomrule
  \end{tabular}
    }
    }

\end{table*}

\begin{table*}[h]
  \vspace{-2em}
  \setlength{\tabcolsep}{4pt}
  \renewcommand{\arraystretch}{1.0}
  \centering
  \caption{Results of different MoEs Configurations}
  \label{tab:results_appendix_more_MoEs_config}
  \resizebox{0.7\textwidth}{!}{
  \begin{tabular}{l|ccccc}
  \toprule
  \textbf{Algorithm} & \textbf{MRPC} & \textbf{CoLA} & \textbf{RTE} & \textbf{STSB} & \textbf{Avg} \\
  \midrule
  standard & 86.83$\scalebox{0.8}{$\pm$0.87}$ & 60.88$\scalebox{0.8}{$\pm$2.54}$ & 78.70$\scalebox{0.8}{$\pm$0.59}$ & 89.07$\scalebox{0.8}{$\pm$0.11}$ & 78.87 \\
  EMoE & 88.05$\scalebox{0.8}{$\pm$0.35}$ & 63.11$\scalebox{0.8}{$\pm$0.21}$ & 80.02$\scalebox{0.8}{$\pm$0.34}$ & 89.37$\scalebox{0.8}{$\pm$0.24}$ & 80.14 \\
  EMoE-learn & 87.93$\scalebox{0.8}{$\pm$0.61}$ & 62.87$\scalebox{0.8}{$\pm$0.71}$ & 79.90$\scalebox{0.8}{$\pm$0.61}$ & 89.40$\scalebox{0.8}{$\pm$0.08}$ & 80.03 \\
  $\scalebox{0.8}{EMoE-last-every2}$ & 87.27$\scalebox{0.8}{$\pm$0.47}$ & 61.60$\scalebox{0.8}{$\pm$0.63}$ & 79.18$\scalebox{0.8}{$\pm$0.17}$ & 89.38$\scalebox{0.8}{$\pm$0.24}$ & 79.36 \\
  $\scalebox{0.7}{EMoE-learn-learn-every2}$ & 87.26$\scalebox{0.8}{$\pm$0.21}$ & 61.82$\scalebox{0.7}{$\pm$1.10}$ & 78.46$\scalebox{0.8}{$\pm$1.22}$ & 89.31$\scalebox{0.8}{$\pm$0.15}$ & 79.21 \\
  $\scalebox{0.7}{EMoE-every2}$ & 86.78$\scalebox{0.8}{$\pm$0.34}$ & 59.21$\scalebox{0.8}{$\pm$0.79}$ & 77.38$\scalebox{0.8}{$\pm$1.04}$ & 89.31$\scalebox{0.8}{$\pm$0.06}$ & 78.17 \\
  $\scalebox{0.7}{EMoE-learn-every2}$ & 86.71$\scalebox{0.8}{$\pm$1.32}$ & 54.02$\scalebox{0.8}{$\pm$0.47}$ & 74.25$\scalebox{0.8}{$\pm$0.74}$ & 88.51$\scalebox{0.8}{$\pm$0.31}$ & 75.87 \\
  \bottomrule
  \end{tabular}
  }
\end{table*}

\begin{table*}[h]
  \setlength{\tabcolsep}{4pt}
  \renewcommand{\arraystretch}{1.0}
  \centering
  \caption{Comparison of EMoE and Standard Results with Different Training Data Fraction}
  \label{tab:results_comparison_datafraction}
  \resizebox{\textwidth}{!}{
  \begin{tabular}{l|ccccccccccccc}
  \toprule
  \textbf{Data} & \textbf{CoLA} & & \textbf{MRPC} & & \textbf{RTE} & & \textbf{STSB} & & \textbf{SST2} & & \textbf{QNLI} & & \textbf{Average} \\ \textbf{Fraction}
  & \textbf{EMoE} & \textbf{Standard} & \textbf{EMoE} & \textbf{Standard} & \textbf{EMoE} & \textbf{Standard} & \textbf{EMoE} & \textbf{Standard} & \textbf{EMoE} & \textbf{Standard} & \textbf{EMoE} & \textbf{Standard} & \textbf{Diff}\\
  \midrule
  1.0 & 62.27 & 60.88 & 87.75 & 86.83 & 80.02 & 78.70 & 89.37 & 89.07 & 95.41 & 95.18 & 92.10 & 91.84 & 0.74 \\
  0.9 & 61.58 & 60.01 & 87.52 & 86.49 & 79.87 & 77.85 & 89.18 & 89.07 & 95.41 & 95.16 & 92.05 & 91.94 & 0.85 \\
  0.8 & 60.89 & 59.28 & 86.58 & 86.35 & 79.22 & 76.77 & 88.98 & 88.99 & 95.34 & 95.06 & 91.94 & 91.83 & 0.78 \\
  0.7 & 59.29 & 58.25 & 86.10 & 85.56 & 77.98 & 76.77 & 87.95 & 87.95 & 95.41 & 95.06 & 92.04 & 91.74 & 0.57 \\
  0.6 & 58.91 & 57.93 & 85.61 & 84.76 & 76.29 & 75.45 & 86.70 & 86.17 & 95.26 & 95.03 & 91.83 & 91.54 & 0.62 \\
  0.5 & 55.18 & 53.89 & 84.91 & 84.76 & 76.53 & 75.21 & 85.63 & 85.59 & 95.19 & 94.91 & 91.23 & 91.12 & 0.53 \\
  0.3 & 50.17 & 50.29 & 82.80 & 82.59 & 73.52 & 72.44 & 80.23 & 79.08 & 94.82 & 94.72 & 90.45 & 90.26 & 0.44 \\
  0.1 & 46.47 & 45.54 & 78.17 & 77.85 & 63.05 & 63.17 & 62.33 & 60.81 & 94.49 & 94.38 & 88.39 & 88.32 & 0.47 \\
  \bottomrule
  \end{tabular}
  }
\end{table*}

\begin{table*}[h]
  \centering
  \caption{Results of top and bottom selection strategies on SST2 and CoLA datasets with different activation ratios. The activation ratio is determined by calculating the proportion of activated neurons that belong to the selected expert among all activated neurons. Meanwhile, the weighted activation ratio is computed by taking the ratio of the sum of activation scores in the selected experts to the sum of the activation scores across the entire FFNs.}
  \label{tab:activation_ratio}
  \resizebox{0.8\textwidth}{!}{
  \begin{tabular}{c|c|cc|cc}
    \toprule
    & & \multicolumn{2}{c|}{Top selection} & \multicolumn{2}{c}{Bottom selection} \\
    Dataset & Activation Ratio & EMoE & EMoE-learn & EMoE & EMoE-learn \\
    \midrule
    \multicolumn{6}{c}{SST2} \\
    \midrule
    \multirow{3}{*}{Activation Ratio} 
    & 32 & 0.6879 & 0.6796 & 0.3121 & 0.0552 \\
    & 16 & 0.4317 & 0.4247 & 0.124 & 0.1293 \\
    & 8 & 0.2616 & 0.2558 & 0.0528 & 0.3218 \\
    \midrule
    \multirow{2}{*}{Weighted Activation Ratio} 
    & 32 & 0.731 & 0.7234 & 0.269 & 0.2791 \\
    & 16 & 0.4953 & 0.4888 & 0.1041 & 0.109 \\
    & 8 & 0.3304 & 0.3237 & 0.0442 & 0.046 \\
    \midrule
    \multicolumn{6}{c}{CoLA} \\
    \midrule
    \multirow{3}{*}{Activation Ratio} 
    & 32 & 0.6815 & 0.6788 & 0.3185 & 0.3212 \\
    & 16 & 0.4292 & 0.4239 & 0.1295 & 0.131 \\
    & 8 & 0.264 & 0.2582 & 0.0552 & 0.0637 \\
    \midrule
    \multirow{2}{*}{Weighted Activation Ratio} 
    & 32 & 0.7341 & 0.4958 & 0.2659 & 0.2744 \\
    & 16 & 0.512 & 0.3389 & 0.0468 & 0.1109 \\
    & 8 & 0.3589 & 0.7256 & 0.1063 & 0.0488 \\
    \bottomrule
  \end{tabular}
  }
\end{table*}

\begin{table*}[h]
  \centering
  \caption{Comparing EMoE and pruning at different states. T-P means training EMoE, pruning experts with lower selection frequency, and evaluating the pruned model. P-T means pruning lower selection frequency experts of an untrained model, then training and evaluating the pruned model. }
  \label{tab:pruning_compare}
  \resizebox{1\columnwidth}{!}
  {
  \begin{tabular}{c|cc|cc}
    \toprule
    & \multicolumn{2}{c}{CoLA}  & \multicolumn{2}{c}{SST2}  \\
    \midrule
    LoRA & \multicolumn{2}{c}{63.40}  & \multicolumn{2}{c}{93.16} \\
    EMoE &\multicolumn{2}{c}{65.33}  & \multicolumn{2}{c}{93.54}\\
    \midrule
    Remained Expert & T-P & P-T & T-P & P-T\\
    \midrule
    64 & 65.26 & 63.4 & 93.42 & 93.16 \\
    32 & 65.33 & 63.52 & 93.54 &93.34 \\
    16 & 65.33 & 63.42 & 93.45 &93.27 \\
    8 & 64.33 & 63.42 & 93.45 & 93.27 \\
    4 & 63.80 & 63.21& 93.34 & 93.21 \\
    \bottomrule
  \end{tabular}
  }
\end{table*}

\section{More Visualization Results}
\label{appdendix:more_visualization}

\begin{figure*}[ht]
\centering
\begin{subfigure}
  \centering
  \includegraphics[width=1.0\linewidth]{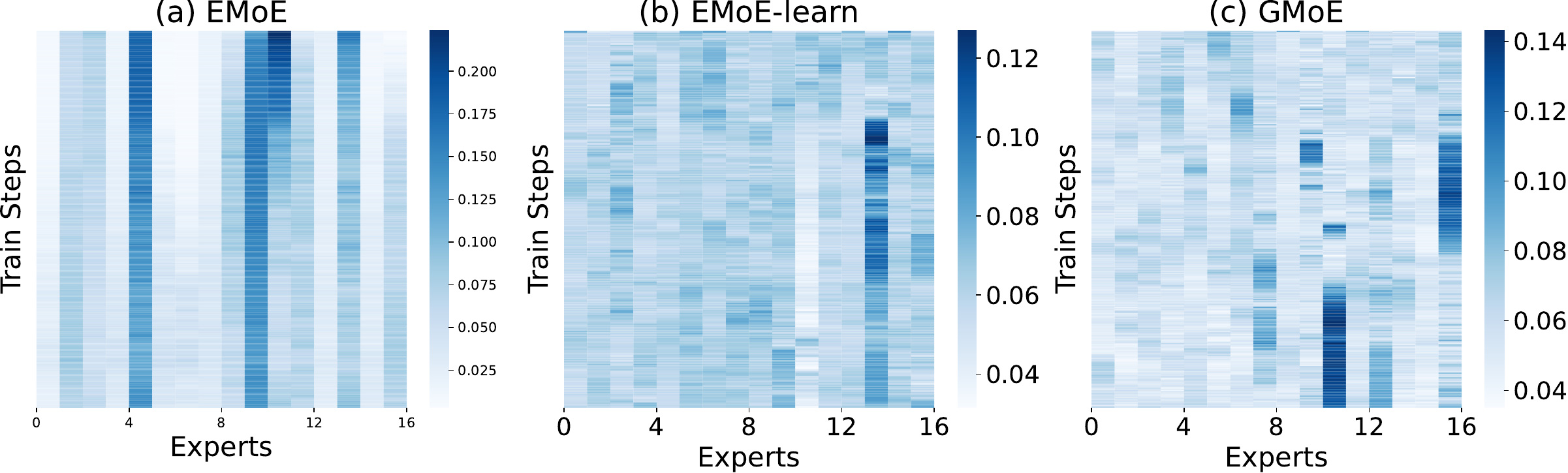}
\end{subfigure}
\begin{subfigure}
  \centering
  \includegraphics[width=1.0\linewidth]{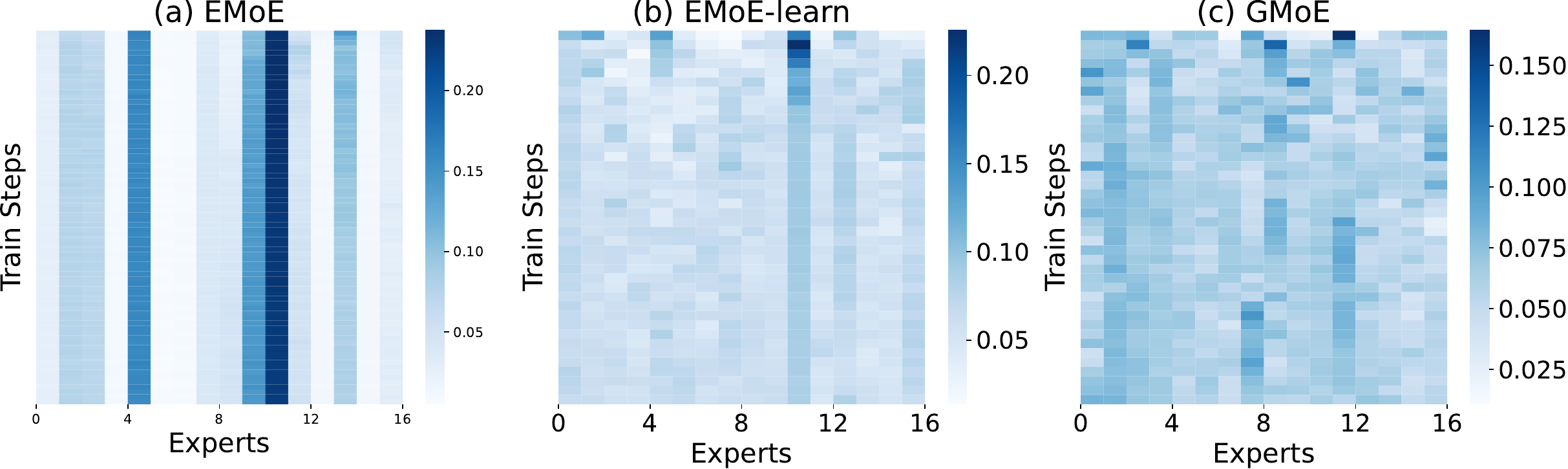}
\end{subfigure}
\begin{subfigure}
  \centering
  \includegraphics[width=1.0\linewidth]{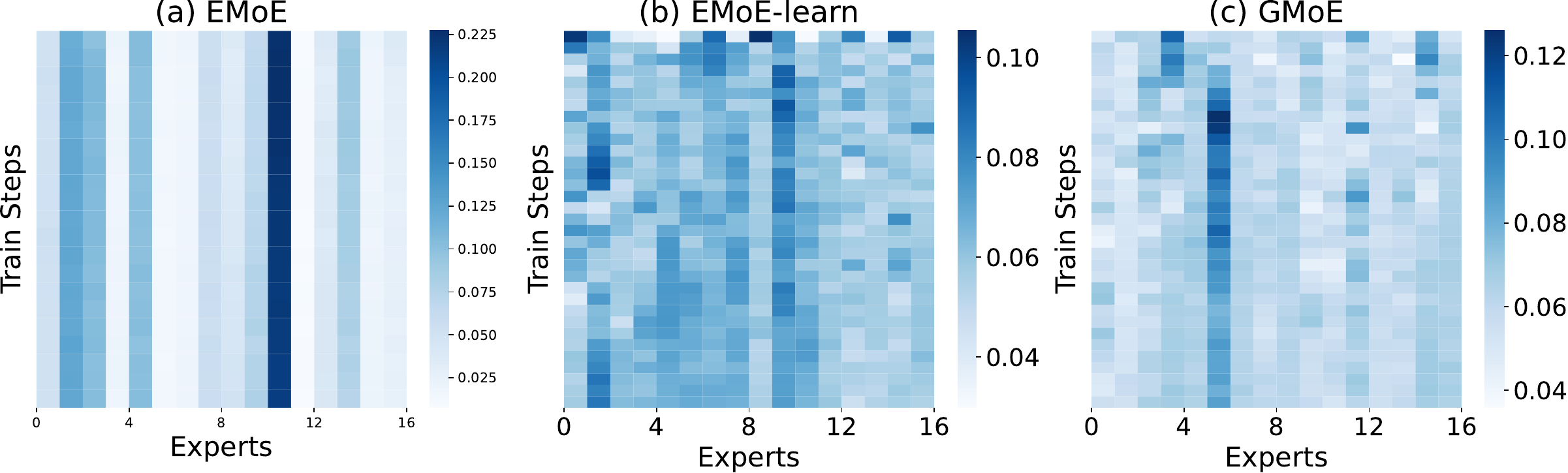}
\end{subfigure}
\begin{subfigure}
  \centering
  \includegraphics[width=1.0\linewidth]{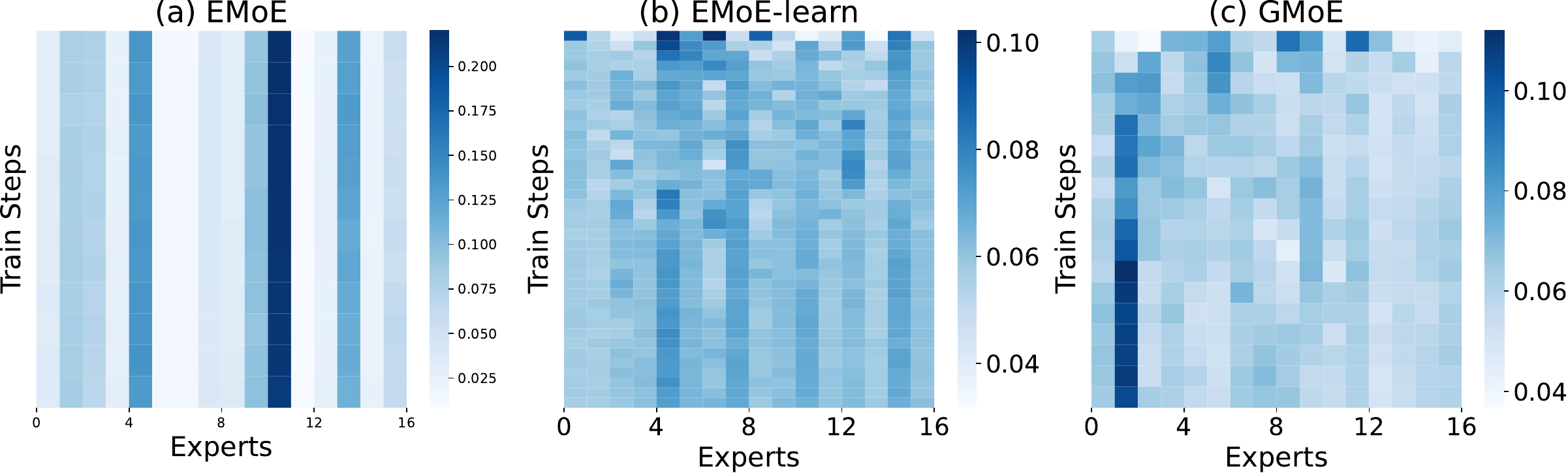}
\end{subfigure}
\caption{Expert selections during training with distinct gating functions ($\operatorname{avg-k}$ vs learned gate) and expert types (modules from the standard vs repetitions of the standard).
The vertical axis illustrates training steps, with top-down arrangement signifying begin-end; the horizontal axis represents expert selection frequency within 1K steps.
(a), (b) and (c) correspond respectively to EMoE, EMoE-learn, and GMoE configurations. The subplots from top to bottom are results for SST-2, STS-B, MRPC, and RTE.}
\label{compare-gates-full}
\end{figure*}

In this part, we demonstrate more gating visualization results on SST-2, STSB, MRPC, and RTE in Figure~\ref{compare-gates-full}. These results are consistent with earlier findings:
(1) Both $\operatorname{avg-k}$ gating and learned gating converge, as indicated by the lower halves of the plots.
(2) $\operatorname{avg-k}$ gating is more stable than learned gating. This could mitigate data inefficiency resulting from inconsistencies in gating across different stages of training~\citep{DBLP:conf/iclr/Zuo00KHZGZ22}.

\end{document}